\documentclass[11pt,a4paper]{article}

\usepackage[hyperref]{emnlp2020}

\usepackage{times}
\usepackage{latexsym}
\usepackage{xspace}
\usepackage[normalem]{ulem}

\newcommand{\EOS}{\texttt{EOS}\xspace}
\newcommand{\EOSp}{\texttt{+EOS}\xspace}
\newcommand{\EOSpOracle}{\texttt{+EOS+Oracle}\xspace}
\newcommand{\EOSm}{\texttt{-EOS}\xspace}
\newcommand{\EOSmOracle}{\texttt{-EOS+Oracle}\xspace}
\newcommand{\dyck}[2]{\mbox{$\text{Dyck-}(#1\text{,}#2$)}\xspace}
\newcommand{\lcutoff}{\mathcal{L}}

\usepackage{microtype}
\usepackage{booktabs, amsmath, amssymb}%
\usepackage{grffile}
\usepackage{graphicx}
\usepackage{newfloat}
\usepackage{makecell}
\usepackage{multicol}
\usepackage{multirow}
\usepackage{subcaption}

\aclfinalcopy %

\newcommand{\AnD}{\hskip 2em plus 1fil minus 0.5em}

\title{The EOS Decision and Length Extrapolation} %
  
 \author{Benjamin Newman \AnD John Hewitt \AnD Percy Liang \AnD Christopher D. Manning \\
 Stanford University \\
 \{\texttt{blnewman,johnhew,pliang,manning\}@stanford.edu}}

\date{}

\begin{document}
\maketitle

\begin{abstract}

Extrapolation to unseen sequence lengths is a challenge for neural generative models of language. %
In this work, we characterize the effect on length extrapolation of a modeling decision often overlooked: predicting the end of the generative process through the use of a special end-of-sequence (\EOS) vocabulary item.
We study an oracle setting---forcing models to generate to the correct sequence length at test time---to compare the length-extrapolative behavior of networks trained to predict \EOS (\texttt{+EOS}) with networks not trained to (\texttt{-EOS}). %
We find that \texttt{-EOS} substantially outperforms \texttt{+EOS}, for example %
extrapolating well to lengths $10$ times longer than those seen at training time in a bracket closing task, %
as well as achieving a $40\%$ improvement over \texttt{+EOS} in the difficult SCAN dataset length generalization task.
By comparing the hidden states and dynamics of \EOSm and \EOSp models, we observe that \EOSp models fail to generalize because they (1) unnecessarily stratify their hidden states by their linear position is a sequence (structures we call \textit{length manifolds}) or (2) get stuck in clusters (which we refer to as \textit{length attractors}) once the \EOS token is the highest-probability prediction.
\end{abstract}

\section{Introduction}

\begin{figure}[!ht]
    \centering
    \includegraphics[width=.48\linewidth]{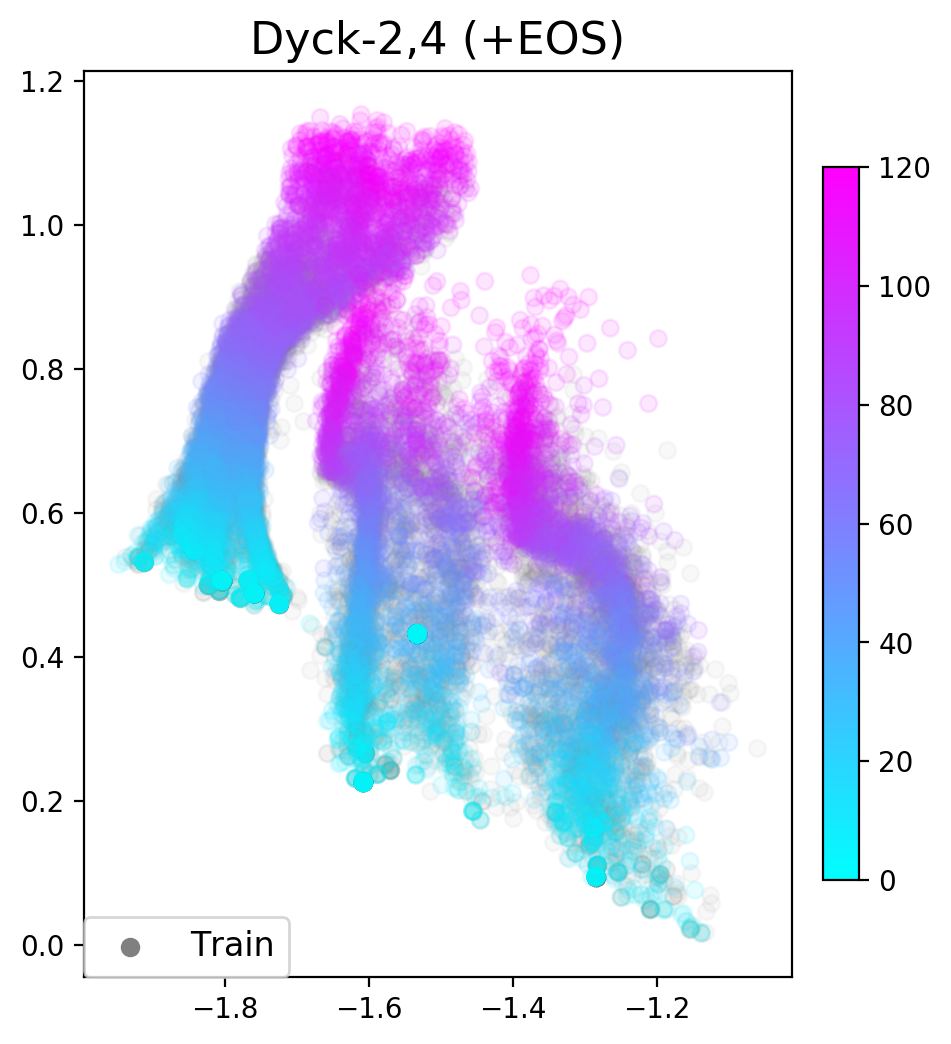}
    \includegraphics[width=.48\linewidth]{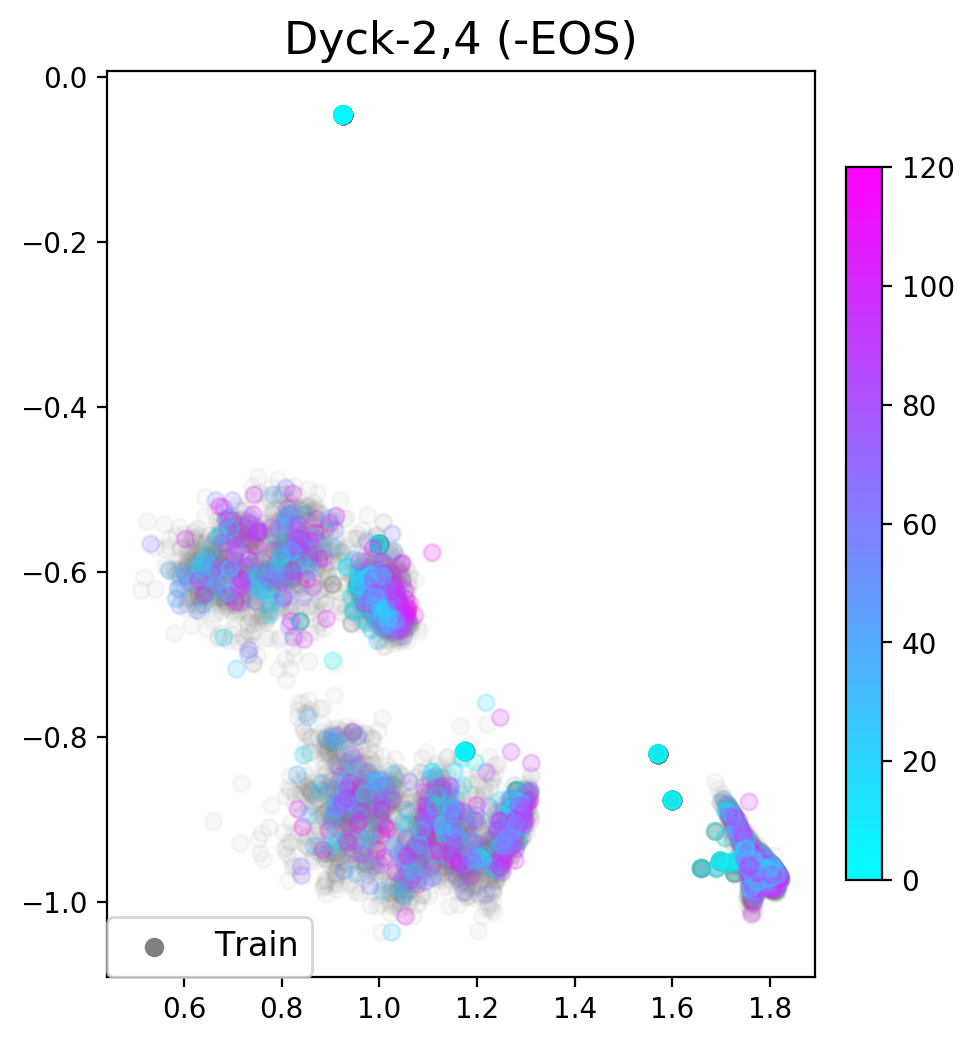}
    \caption{The hidden state dynamics differ between a model that has to predict \EOS ({\bf left}) and one that does not ({\bf right}). Color varies with the hidden states' position in the sequence.}
    \label{fig:state_b}
\end{figure}

A core feature of the human language capacity is the ability to comprehend utterances of potentially unbounded length by understanding their constituents and how they combine \citep{frege1953foundations,chomsky1957syntactic,montague1970universal}.
In NLP, while better modeling techniques have improved models' abilities to perform some kinds of systematic generalization \citep{gordon2019permutation, lake2019compositional}, these models still struggle to extrapolate; they have trouble producing and processing sequences longer than those seen during training even if they are composed of familiar atomic units.

In this work, we investigate how an understudied modeling decision affects the ability of neural generative models of language to extrapolate to longer sequences. %
Models need to place a distribution over all sequence lengths, and to accomplish this when they generate from left to right, they make a decision about whether to generate a content token or a special end-of-sequence (\EOS) token at each position in the sequence.
In practice, however, the decision to have models predict where sequences end in this way, which we refer to as the \textit{\EOS decision}, yields models that predict sequences should end too early in length-extrapolation settings \cite{hupkes2020compositionality, dubois2019location}. %

We conceptually decompose length extrapolation into two components: (i) the ability to produce the right content, and (ii) the ability to know when to end the sequence.
Standard evaluations like exact match or BLEU  \cite{papineni-etal-2002-bleu} on length-generalization datasets like SCAN and gSCAN \cite{lake2018generalization,ruis2020benchmark} evaluate the ability to perform both components, but models are known to fail because of (ii) alone, so these metrics don't test (i) in isolation. %
To help evaluate (i) independent of (ii), another evaluation tool has been to exclude the effect of the \EOS decision at test time by forcing models to generate to the correct output length \cite{lake2018generalization} or evaluating whether the strings they generate match prefixes of the correct outputs \cite{dubois2019location, hupkes2020compositionality}. %

In contrast, our work is the first to explore what happens if we don't have to perform (ii) (predicting where to end the sequence) \textbf{even at training time}.
In this case, can we perform better at (i) (generating the right content)? %
We endeavor to answer this question by comparing the extrapolative ability of models trained without \EOS tokens (\EOSm) to models trained with them in their training data (\EOSp).

First, we look at a simple formal language called \dyck{k}{m} that contains strings of balanced parentheses of $k$ types with a bounded nesting depth of $m$ \cite{hewitt2020dyckkm}; see Table~\ref{tab:dyck-sample}.
Looking at a simple bracket closing task we find that \EOSm models close brackets with high accuracy %
on sequences 10 times longer than those seen at training time, whereas \EOSp models perform substantially worse, suggesting fundamental differences in these models' ability to extrapolate.
Investigating models' hidden states under principal component analysis (PCA) reveals that \EOSp models unnecessarily stratify their hidden states by linear position while \EOSm models do so less, likely contributing to \EOSm models' improved performance (Figure \ref{fig:state_b}). We call these stratified hidden states \textit{length manifolds}.

Second, we use SCAN---a synthetic sequence-to-sequence dataset meant to assess systematic generalization.
We focus on the split of the dataset made to test length extrapolation, showing that models in the \EOSp condition perform up to 40\% worse than those in the \EOSm condition in an oracle evaluation setting where models are provided the correct output length.
Visualizing the hidden states of both models using PCA reveals that \EOSp models fail once they put high probability on the \EOS token because their hidden states remain in place, exhibiting what we term a \textit{length attractor}.
\EOSm models do not predict the \EOS token, and their hidden states are able to move throughout hidden state space during the entire generation process, likely contributing to their improved performance.

Finally, we investigate extrapolation in a human language task---translation from German into English.
We use an oracle evaluation where sequence lengths are chosen to maximize the BLEU score of generated translations.
In contrast to our other experiments, for this more complex NLP task, we see that the \EOSm condition less consistently contributes to extrapolation ability.

\section{Related Work}

\paragraph{The Difficulty of Extrapolation.}
Evidence from a number of settings shows that processing sequences longer than those seen at training time is a challenge for neural models \citep{dubois2019location, ruis2020benchmark, hupkes2020compositionality, klinger2020study}.
This difficulty has been observed in datasets designed to test the ability of models to compositionally generalize, such as SCAN \citep{lake2018generalization}, where the best performing neural models do not even exceed 20\% accuracy on generating sequences of out-of-domain lengths, whereas in-domain performance is 100\%. %
Extrapolation has also been a challenge for neural machine translation; \citet{murray2018correcting} identifies models producing translations that are too short as one of the main challenges for neural MT.

There are a number of reasons why extrapolation is challenging.
At the highest level, some believe extrapolation requires understanding the global structure of a task, which standard neural architectures may have trouble emulating from a modest number of in-domain samples \citep{mitchell2018extrapolation, marcus2018deep, gordon2019permutation}.
Others focus more on implementation-level modeling problems, such as the \EOS problem, where models tend to predict sequences should end too early, and try to address it with architectural changes \citep{dubois2019location}. %
Our work focuses on the latter issue; we show how the \EOS decision affects not just a model's tendency for producing \EOS tokens too early, but also the representations and dynamics underlying these decisions.

\paragraph{Addressing Extrapolation Issues}
Previous approaches to address the poor extrapolation ability of standard sequence models often focus on architectural changes. 
They include incorporating explicit memory \citep{graves2014neural} or specialized attention mechanisms \citep{dubois2019location}.
Others have proposed search-based approaches---modifying beam search to prevent \EOS from being assigned too much probability mass \citep{murray2018correcting}, or searching through programs to produce sequences rather than having neural models produce sequences themselves \citep{Nye2020LearningCR}.
Extrapolative performance has recently come up in the design of positional embeddings for Transformers \cite{vaswani2017attention}.
While recent Transformer models like BERT use learned positional embeddings \citep{devlin2018bert}, when they were first introduced, position was represented through a combination of sinusoids of different periods.
These sinusoids were introduced with the hope that their periodicity would allow for better length extrapolation \cite{vaswani2017attention}.
In this study we provide insight into popular architectures and the \EOS decision instead of pursuing new architectures. %

\paragraph{Length in Sequence Models}
Historically, for discrete symbol probabilistic sequence models, no \EOS token was included in the standard presentations of HMMs \citep{rabiner1989tutorial} or in most following work in speech. 
However, the importance of having an \EOS token in the event space of probabilistic sequence models was emphasized by NLP researchers in the late 1990s (e.g. \citet{collins1999head} Section 2.4.1) and have been used ever since, including in neural sequence models.
Requiring neural models to predict these \EOS tokens means that these models' representations must track sequence length.
There is empirical evidence that the hidden states of LSTM sequence-to-sequence models trained to perform machine translation models track sequence length by implementing something akin to a counter that increments during encoding and decrements during decoding \citep{shi2016neural}.
These results are consistent with theoretical and empirical findings that show that LSTMs can efficiently implement counting mechanisms \citep{weiss2018practical, suzgun2019lstm, merrill2020linguistic}.
 Our experiments will show that tracking absolute token position by implementing something akin to these counters makes extrapolation difficult. %

\section{Experimental Setup}
In all of our experiments, we compare models trained under two conditions: \EOSp, where \EOS tokens are appended to each training example, as is standard; and \EOSm, where no \EOS tokens are appended to training examples.

For our \dyck{k}{m} experiments (Section \ref{section:dyck}), we train on sequences with lengths from a set $\lcutoff_{\text{train}}$ and test on sequences with lengths from a set $\lcutoff_{\text{test}}$, where $\max \lcutoff_{\text{train}} < \min \lcutoff_{\text{test}}$.

For our SCAN and German-English MT experiments (Sections \ref{section:scan} and \ref{section:mt}), we train on sequences of tokens shorter than some length $\ell$, and test on sequences longer than $\ell$.
We refer to $\ell$ as a \textbf{length cut-off}, and a train-test split created in this manner is referred to as a \textbf{length-split}.
We evaluate these models differently in each experiment.

Because we are interested in assessing the extrapolative ability of models apart from predicting where sequences end, we use what we refer to as a \textbf{length oracle evaluation}.
It is performed by first allowing \EOSm and \EOSp models to output sequences longer than any gold targets, and then choosing the best position to end the sequence.
For the \EOSp models, this requires preventing the models from emitting \EOS tokens during generation by setting the probability mass on the \EOS token to $0$.
We call these models \EOSmOracle and \EOSpOracle respectively.
Using these oracle metrics ensures that models are only assessed on knowing what content to generate rather than where to end sequences.

\begin{figure}
    \centering
    \includegraphics[width=0.8\linewidth]{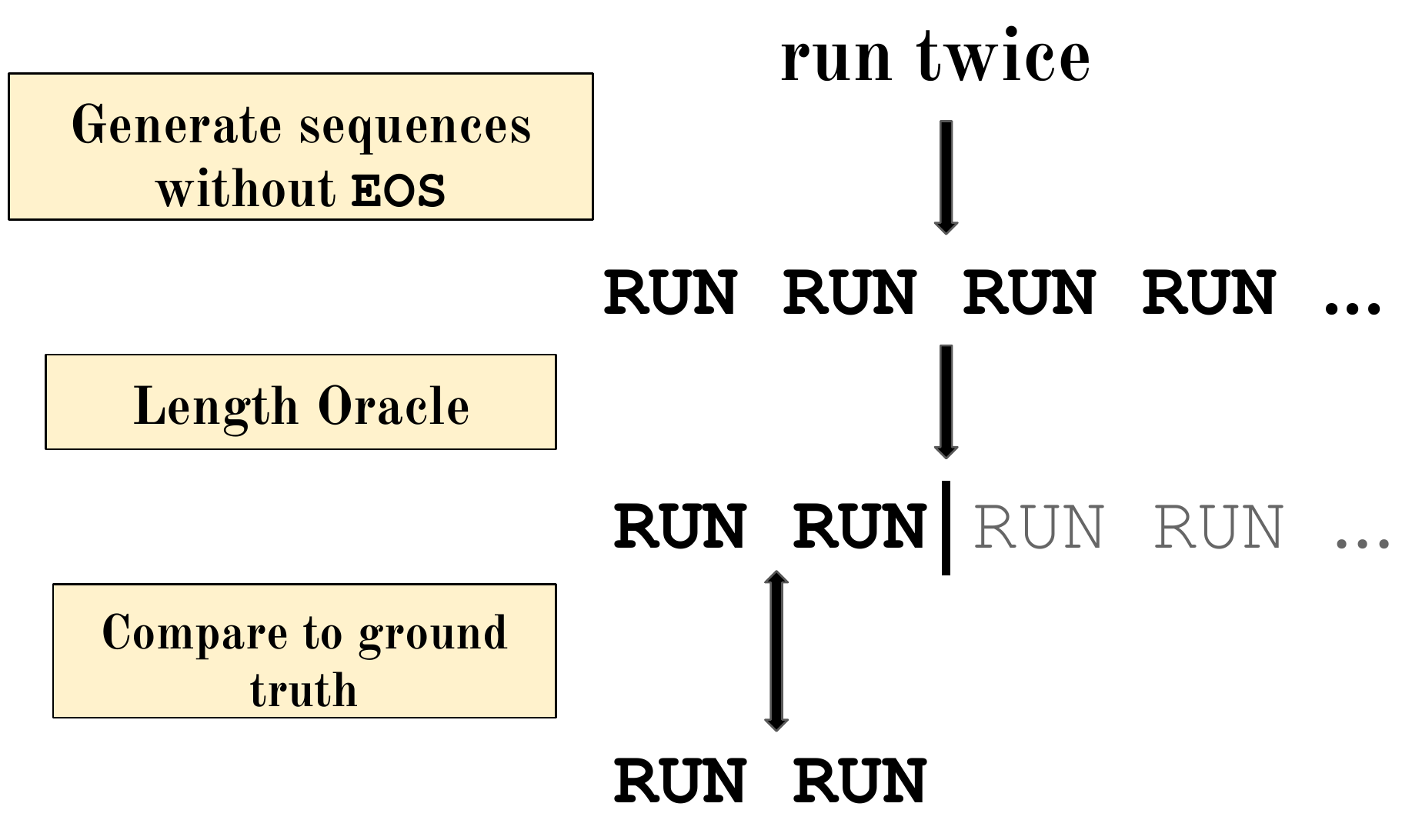}
    \caption{Above is an example of how \textbf{length oracle evaluation} works with the SCAN task. Models (both \EOSp and \EOSm) generate sequences without producing \EOS tokens. The length oracle decides the optimal place to end the sequences, and compares the result to the ground truth.}
    \label{fig:length_oracle_example}
\end{figure}

\section{Experiment 1: \dyck{k}{m}}
\label{section:dyck}
We conduct our first experiments with a simple language that gives us the opportunity to train an accurate, interpretable language model.
This allows us to develop an intuition for how the \EOS token affects the representations that models learn.

The language we study is \dyck{k}{m} \citep{hewitt2020dyckkm}. %
It is a formal language that consists of strings of well-nested, balanced brackets.
There are $k$ types of brackets, and a maximum nesting depth of $m$.
For our experiments, we set $k=2$, and vary $m$.
Accepting this language involves implementing a bounded stack of up to size $m$, and   allowing for pushing and popping any of $k$ distinct elements (the bracket types) to and from the stack.
We use the term \textbf{stack state} to refer to what this stack looks like at a given point in processing a sequence. An example sequence and its stack states are given in Table~\ref{tab:dyck-sample}.

\begin{table}
    \centering
    \small
    \renewcommand{\tabcolsep}{2pt}
    \begin{tabular}{c c c c c c c c c c c c c}
    \toprule
         sequence: &  ( & [ & ( & ) & [ & [ & ( & ) & ] & ] & ] & ) \\
         stack state: & ( & ([ & ([( & ([ & ([[ & ([[[ & ([[[( & ([[[ & ([[ & ([ & ( & \\ 
    \bottomrule
    \end{tabular}
    \caption{A sample sequence and its stack states from \dyck{2}{6} and \dyck{2}{8}, but not \dyck{2}{4} because the maximum nesting depth is 5.}
    \label{tab:dyck-sample}
\end{table}

\subsection{Methodology}
We train our \EOSp and \EOSm models on a language modeling task using samples from \dyck{2}{4}; \dyck{2}{6}; and \dyck{2}{8}.
Minimum training sample lengths are chosen such that sequences of that length contain stack states of all possible depths at least three times in expectation.
We include enough training samples to reliably achieve perfect bracket-closing accuracy on an in-domain evaluation.\footnote{The code to generate these samples is available at \url{https://github.com/bnewm0609/eos-decision}.}
Including \textit{only} long sequences of brackets means that models only see evidence that they can end sequences (in the form of an \EOS token) after a long stretch of symbols, not whenever their stack states are empty %
(and the \EOSm model sees no evidence of ending at all).
We endeavor to address this shortcoming in Section \ref{section:dyck-eos}.

The out-of-domain samples (with lengths in $\lcutoff_{\text{test}}$) are $10$ times longer than in-domain ones (with lengths in $\lcutoff_{\text{train}}$). %
(Table~\ref{tab:dycklengths} gives the lengths for each condition.)
The models we use are single layer LSTMs with $5m$ hidden states, and we train each model such that it has perfect in-domain length accuracy on a held-out validation set to ensure that we learn the in-domain data distribution well, rather than sacrifice in-domain performance for potential gains in extrapolative performance.
Additionally, such in-domain accuracy is only possible because we are bounding the stack depth of our \dyck{k}{m} samples.
If we were using the unbounded Dyck-$k$ languages (with $k>1$), we would not be able to see good in-domain performance, and thus we would not expect to see acceptable out-of domain performance either \cite{suzgun2019evaluating,suzgun2019lstm}.

For evaluation, we use the same metric as \citet{hewitt2020dyckkm} as it evaluates the crux of the task: maintaining a long-term stack-like memory of brackets.
In principle, we want to ensure that whenever a model can close a bracket, it puts more probability mass on the correct one.
Our metric is as follows: whenever it is acceptable for the model to close a bracket, we compute how often our models assign an large majority (more than 80\%) of the probability mass assigned to \textit{any} of the closing brackets to the \textit{correct} closing bracket.
For example, if $20\%$ of the probability mass is put on either closing bracket, the model gets credit if it puts more than $16\%$ of the mass on the correct bracket.
Because in each sample there are more bracket pairs that are (sequentially) close together than far, we compute this accuracy separately for each distance between the open and close bracket. %
Our final score is then the mean of all these accuracies.

We also report a standard metric: perplexity for sequences of in-domain and out-of-domain lengths.
We choose perplexity because it incorporates the probabilistic nature of the task---we cannot evaluate using exact match because at almost all decoding steps (other than when the stack has $m$ brackets on it) there are two equally valid predictions that the model can make.

\subsection{Results}
\begin{table}
    \centering
    \small
    \begin{tabular}{c c c c c}
    \toprule
         $m$ & $\lcutoff_{\text{train}}$ & $\lcutoff_{\text{test}}$ & \EOSp & \EOSm \\
         \midrule
         4 & [88, 116]  & [950, 1050]  & 0.60 & \bf0.86\\ %
         6 & [184, 228] & [1840, 2280] & 0.68 & \bf0.98\\ %
         8 & [328, 396] & [3280, 3960] & 0.68 & \bf0.96\\ %
    \bottomrule
    \end{tabular}
    \caption{\dyck{k}{m} bracket closing metric results, median of 5 independent training runs.}
    \label{tab:dycklengths}
\end{table}

\begin{figure*}
\centering
     \includegraphics[width=.23\linewidth]{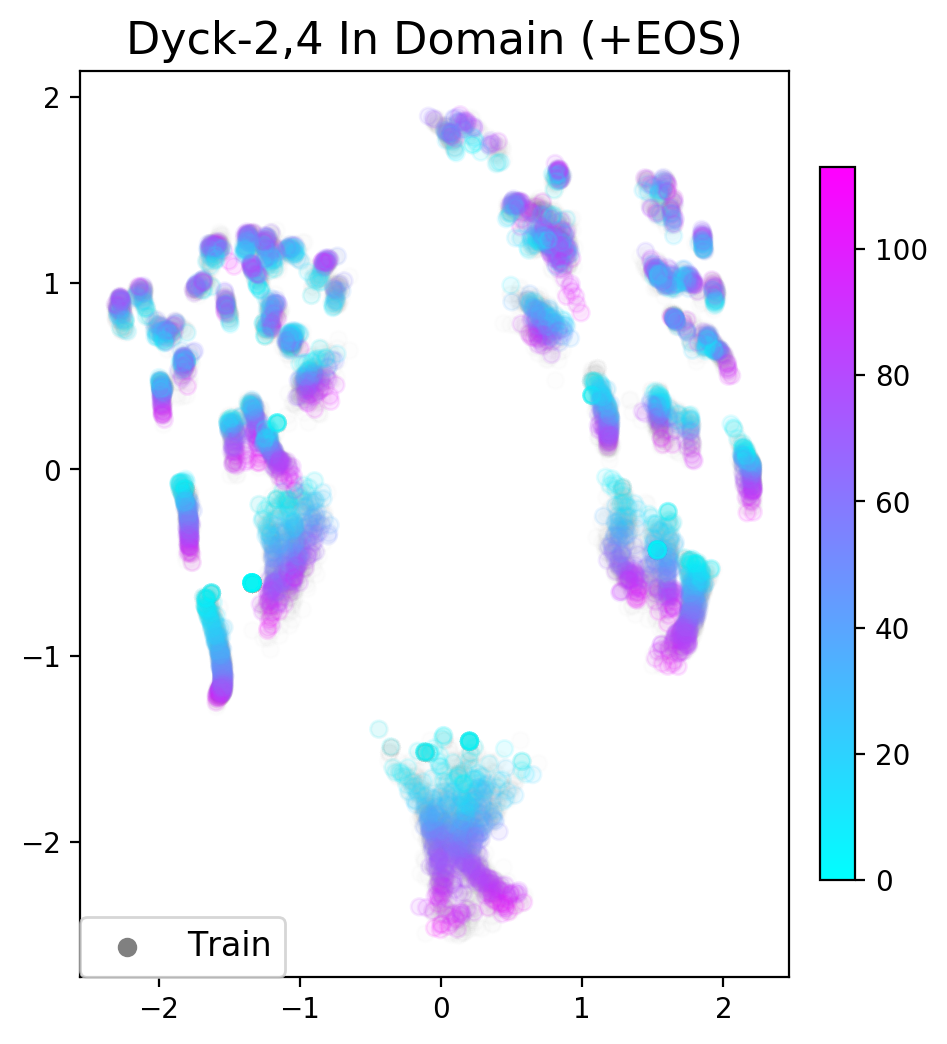} 
     \includegraphics[width=.23\linewidth]{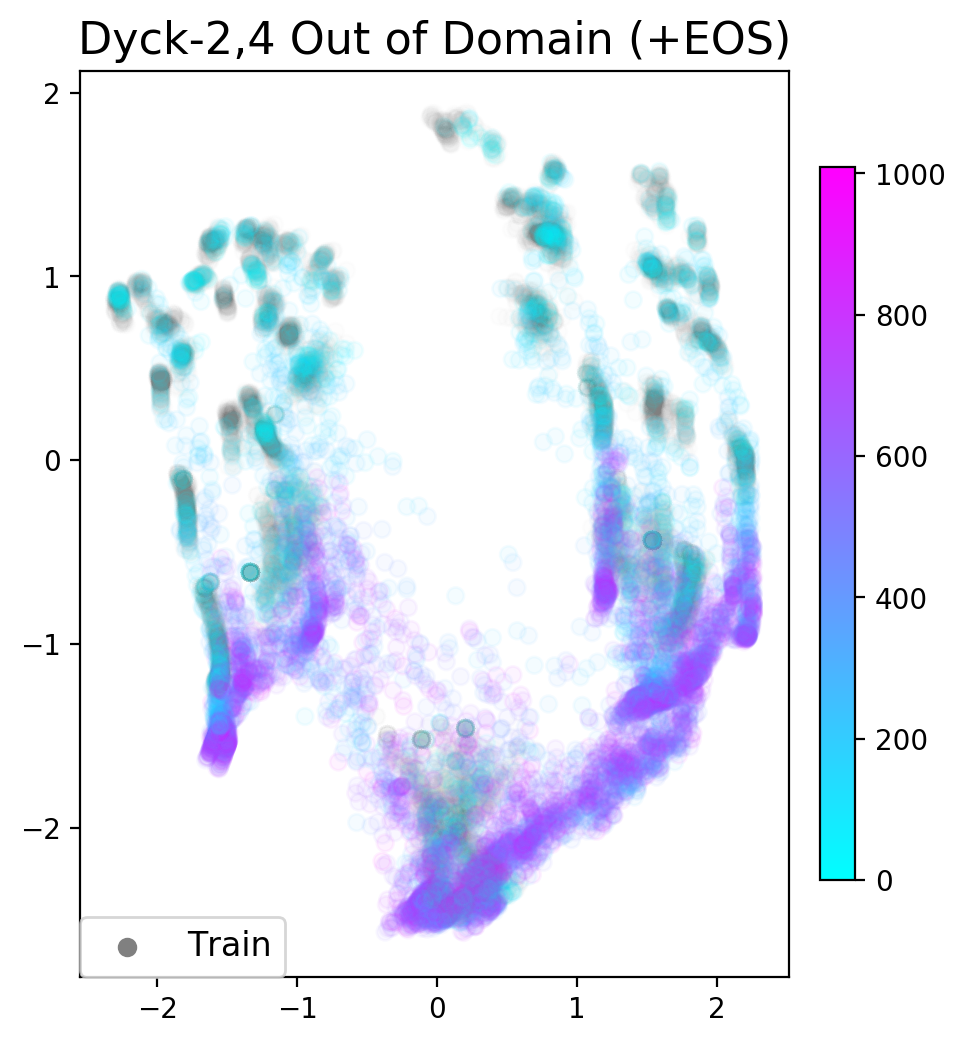}
     \includegraphics[width=.23\linewidth]{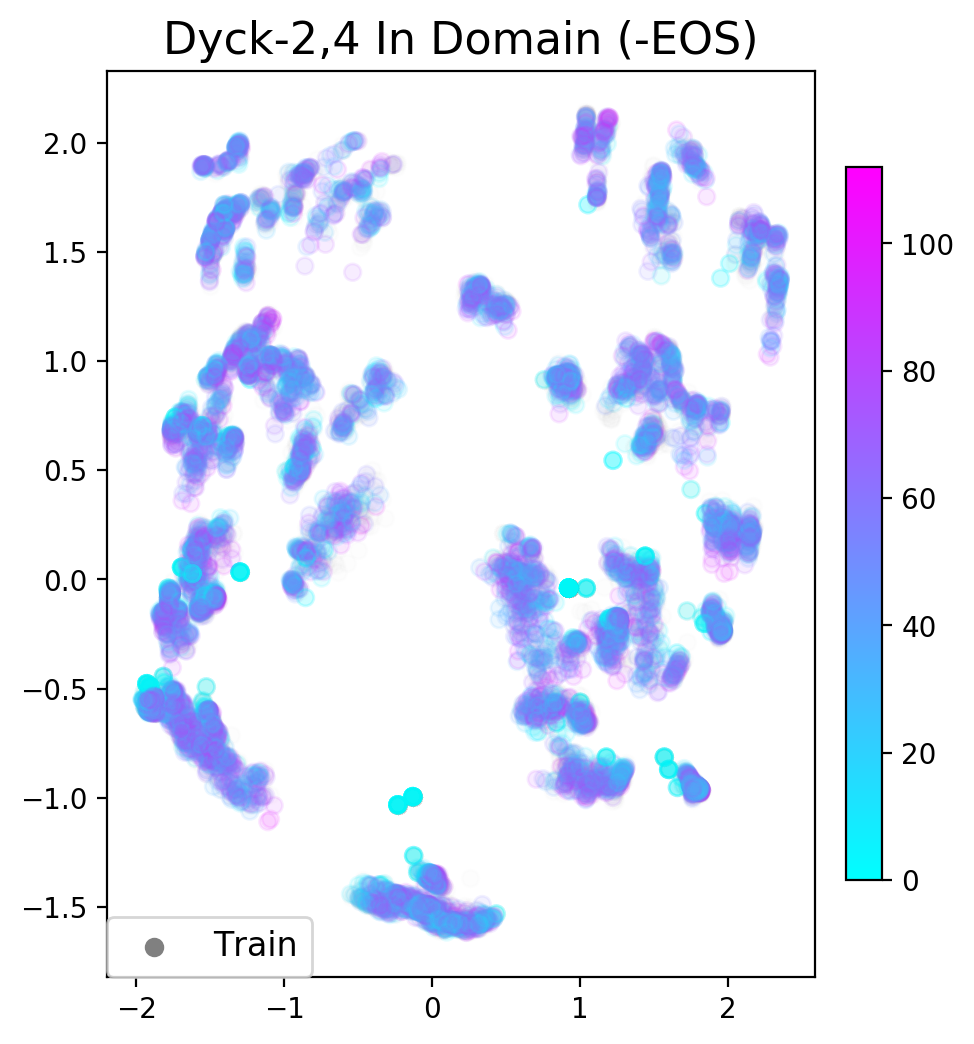}
     \includegraphics[width=.23\linewidth]{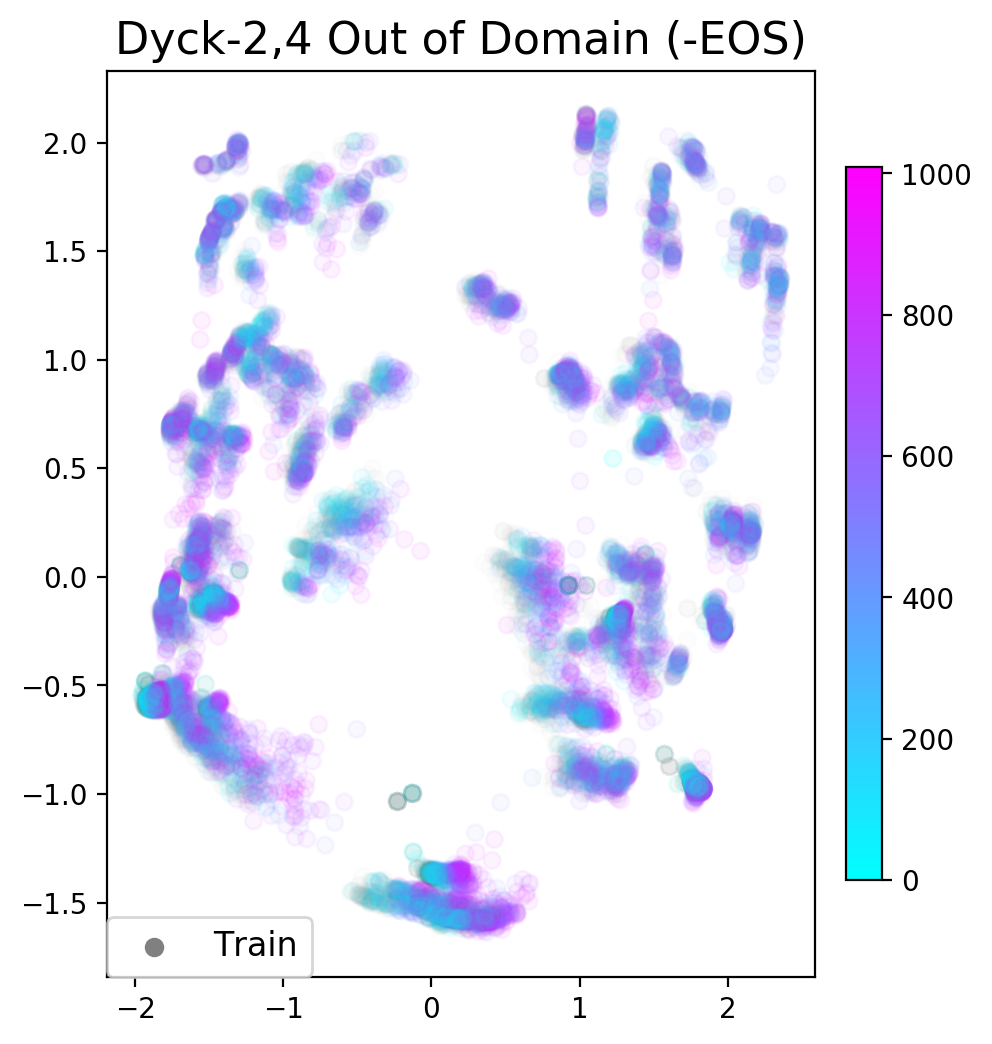}
    \caption{The top two principal components explain $67.52\%$ of the variance for the \EOSp model and $62.27\%$%
    of the variance for the \EOSm hidden states. The color scale corresponds to index in the sequence. The bottom-most cluster (centered around $(0, -2.0)$ in the \EOSp plot and $(0, -1.5)$ in the \EOSm plot) contains all of the hidden states where the stack is empty and the sequence can end.}
    \label{fig:dyck42pca}
\end{figure*}

Recall that we train all models to obtain perfect in-domain bracket closing accuracy on a held-out validation set. However, the \EOSp models' performance is severely degraded on out-of-domain sequences 10 times longer than in-domain ones, while the \EOSm models still perform well across all conditions (Table~\ref{tab:dycklengths}). 
The perplexity results mirror the results of the bracket closing task (Appendix \ref{app:plots-dyck}).

Because the \dyck{k}{m} is relatively simple, the model's hidden states are interpretable. We run the model on the training data, stack the models' hidden states, extract their top two principal components, and note that they form clusters (Figure~\ref{fig:dyck42pca}) that encode stack states (Appendix; Figure~\ref{fig:plots-dyck24}).
We observe that as both models process sequences, they hop between the clusters. However, in the \EOSp model, the hidden states move from one side of each cluster to the other during processing, while this does not happen in the \EOSm model.
Because of this, we refer to these elongated clusters as \textbf{length manifolds}.

As the \EOSp model processes inputs past the maximum length it has seen during training, the representations lie past the boundary of the clusters, which we observe leads to a degradation of the model's ability to transition between states.
We hypothesize this is because the recurrent dynamics break near the edges of the training-time length manifolds.
We can see this degradation of the model dynamics in the out-of-domain \EOSp plot (Figure \ref{fig:dyck42pca}).
The hidden states at positions further in the sequences congregate at the tips of the length manifolds, or completely deviate from the training data.
Contrast this plot to the in-domain \EOSm plot, where there are no length manifolds visible and the out-of-domain \EOSm plot, where we can see some length manifolds, but less evidence of the degradation of predictions (Table~\ref{tab:dycklengths}).

We hypothesize that these length manifolds form because \EOSp models need to predict where each sequence ends in order to confidently put probability mass on the \EOS token.\footnote{Close brackets are more likely than open brackets at training time near the maximum training length, since all sequences must end in the empty stack state; this may explain why \EOSm models still show some length-tracking behavior.}
Our main takeaway is then that this length tracking has negative effects when the model is used to predict longer sequences.

\subsection{Predicting the End of Sequences}
\label{section:dyck-eos}
Knowing where a \dyck{k}{m} sequence can end is an important part of being able to accept strings from the language. While we focus more on the ability of models to extrapolate in this work, and our evaluations do not depend on the ability of models to end sequences, we do observe in our plots of hidden states that the cluster of hidden states corresponding to the empty stack state (where sequences can end) is linearly separable from the other clusters (Figure \ref{fig:dyck42pca}). This linear-separability suggests we can train a linear classifier on top of model hidden states, a probe, to predict where sequences can end \citep{conneau2018cram, ettinger-etal-2018-assessing}. 

We train a binary classifier to predict if a sequence can end at every token position. The input to this classifier at time-step $t$ is the sequence models' hidden state at $t$, and it is trained to predict $1$ if the sequence can end and $0$ otherwise. We find that this simple method is able to predict where sequences end with high accuracy for in-domain sequences, and higher accuracy for the \EOSm models compared to the \EOSp models for the longer out-of-domain sequences (Table~\ref{tab:dyck-probe}). This distinction mirrors the disorganization of the hidden states in the \EOSp models hidden states when they extrapolate (Figure \ref{fig:dyck42pca}).

\begin{table}
    \centering
    \small
    \begin{tabular}{c c c c c}
        \toprule
         m & \multicolumn{2}{c}{\EOSp} & \multicolumn{2}{c}{\EOSm}  \\
         \midrule
           & ID & OOD & ID & OOD \\
         4 & 1.0 & 0.90 & 1.0 & 0.92 \\
         6 & 1.0 & 0.95 & 1.0 & 1.0 \\
         8 & 1.0 & 0.97 & 1.0 & 1.0 \\
         \bottomrule
    \end{tabular}
    \caption{Accuracies for predicting when \dyck{2}{m} sequences can end. Each entry is the median of five runs. Note that even though \EOSp model is trained with an \EOS token, it does not receive signal that it can end at other empty stack states, so such a probe is needed for models to learn when to end sequences.}
    \label{tab:dyck-probe}
\end{table}

\section{Experiment 2: SCAN}
\label{section:scan}

While the \dyck{k}{m} experiments give us good intuitions for how and why \EOS tokens affect extrapolation, we study SCAN to extend to sequence-to-sequence modeling, and since it is a well-studied dataset in the context of length extrapolation.

SCAN is a synthetic sequence-to-sequence dataset meant to simulate data used to train an instruction-following robot \citep{lake2018generalization}.
The inputs are templated instructions while the outputs correspond to the sequences of actions the robot should take. For example, the input instruction ``walk left twice"  maps to the output action sequence \texttt{TURN\_LEFT WALK TURN\_LEFT WALK}. There are 13 types in the input vocabulary---five actions (turn, run, walk, jump, look); six modifiers (left, right, opposite, around, twice, thrice); and two conjunctions (and, after), and different combinations of these tokens map to different numbers of actions. For example, ``turn left twice" maps to \texttt{TURN\_LEFT TURN\_LEFT} while ``turn left thrice'' maps to \texttt{TURN\_LEFT TURN\_LEFT TURN\_LEFT}.
This is important for our exploration of models' abilities to extrapolate because it means that while the lengths of the input instructions range from one to nine tokens, the output action sequences lengths vary from one to 48 tokens.

\subsection{Length splits in SCAN}
\citet{lake2018generalization} define a 22-token length split of the dataset, in which models are trained on instructions with sequences whose outputs have 1--22 tokens (16990 samples) and are evaluated on sequences whose outputs have 22--48 tokens (3920 samples).
Previous work has noted the difficulty of this length split---the best approach where sequences are generated by a neural model achieves $20.8\%$ accuracy \citep{lake2018generalization}.\footnote{Note that \citet{Nye2020LearningCR} are able to achieve $100\%$ accuracy on the SCAN length split, though their set-up is different---their neural model searchers for the SCAN grammar rather than generating sequences by token.} %
Specifically, this extra difficulty stems from the fact that neural models only perform the SCAN task well when they are trained with sufficient examples of certain input sequence templates \citep{loula2018rearranging}.
In particular, the \citet{lake2018generalization} length split lacks a template from the training data: one of the form ``walk/jump/look/run around left/right thrice.'' Alone, this template results in an output sequence of 24 actions, and thus is not included in the training set.
Approximately 80\% of the test set is comprised of sequences with this template, but models have not seen any examples of it at training time, so they perform poorly.
Thus, we create additional length splits that have more examples of this template, and evaluate using them in addition to the standard 22-token length split of SCAN. %

\subsection{Methodology}
We generate the 10 possible length splits of the data with length cutoffs ranging from 24 tokens to 40 (See the header of Table~\ref{tab:scan-results}).
Not every integer is represented because SCAN action sequences do not have all possible lengths in this range.
For each split, we train the LSTM sequence-to-sequence model that \citet{lake2018generalization} found to have the best in-domain performance.
Like in the previous experiment, we train two models---a \EOSp model with an \EOS token appended to each training sample output, and a \EOSm model without them.
We use the standard evaluation of \citet{lake2018generalization}: greedy decoding and evaluating based on exact match.
Because we are interested in how well our models can extrapolate separately from their ability to know when an output should end, we use force our models to generate to the correct sequence length before comparing their predictions to the gold outputs.
For the \EOSp model, this entails preventing the model from predicting \EOS tokens by setting that token's probability to $0$ during decoding.
This gives us the \EOSmOracle and \EOSpOracle evaluations.
We additionally report the exact match accuracy when the \EOSp model is allowed to emit an \EOS token (\EOSp), which is the standard evaluation for this task \citep{lake2018generalization}.

\subsection{Results}
\begin{table*}
    \centering
    \small
    \begin{tabular}{c l r r r r r r r r r r r}
        \toprule
        & $\ell$ \textit{(length cutoff)} & 22 & 24 & 25 & 26 & 27 & 28 & 30 & 32 & 33 & 36 & 40 \\
        \midrule
        \parbox[t]{3mm}{\multirow{3}{*}{\rotatebox[origin=c]{90}{LSTM}}}  & \it \EOSp & \it 0.16 &\it 0.08 &\it 0.26 & \it0.61 & \it0.72 & \it0.64 &\it 0.60 &\it 0.67 &\it 0.45 &\it 0.47 &\it 0.85 \\
        & \EOSpOracle & 0.18 & 0.46 & 0.47 & 0.71 & 0.82 & 0.77 & 0.80 & 0.84 & 0.74 & 0.81 & 0.95 \\
        & \EOSmOracle & \textbf{0.61} & \textbf{0.57} & \textbf{0.54} & \textbf{0.83} & \textbf{0.92} & \textbf{0.97} & \textbf{0.90} & \textbf{1.00} & \textbf{0.99} & \textbf{0.98} & \textbf{1.00} \\
        \midrule
        \parbox[t]{3mm}{\multirow{3}{*}{\rotatebox[origin=c]{90}{\tiny Transformer}}} & \it\EOSp & \it0.00 & \it0.05 & \it0.04 & \it0.00 & \it0.09 & \it0.00 & \it0.09 & \it0.35 & \it0.00 & \it0.00 & \it0.00\\
         & \EOSpOracle & 0.53 & 0.51 & \bf0.69 & 0.76 & 0.74 & 0.57 & 0.78 & 0.66 & 0.77 & \bf1.00 & 0.97\\
         & \EOSmOracle & \bf0.58 & \bf0.54 & 0.67 & \bf0.82 & \bf0.88 & \bf0.85 & \bf0.89 & \bf0.82 & \bf1.00 & \bf1.00 & \bf1.00\\
        \bottomrule
    \end{tabular}
    \caption{Exact match accuracies on length splits. Reported results are the median of 5 runs. The \EOSp rows (italicized) are not comparable to the other rows because they are not evaluated in an an oracle setting---they are provided for reference.}
    \label{tab:scan-results}
\end{table*}

\begin{figure}[!ht]
\centering
    \subcaptionbox{\label{fig:scan-pca-a}}{\includegraphics[width=0.92\linewidth]{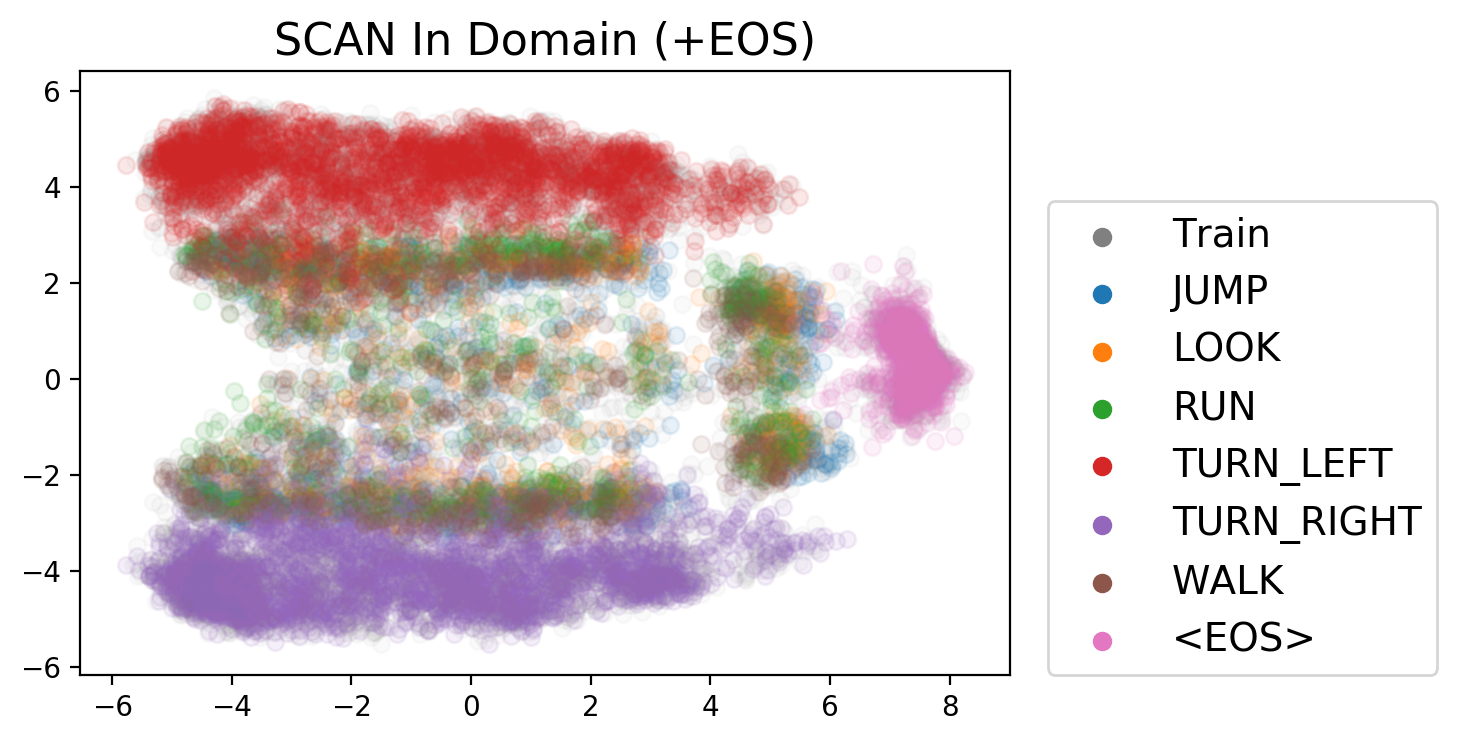}}
    \subcaptionbox{\label{fig:scan-pca-b}}{\includegraphics[width=0.92\linewidth]{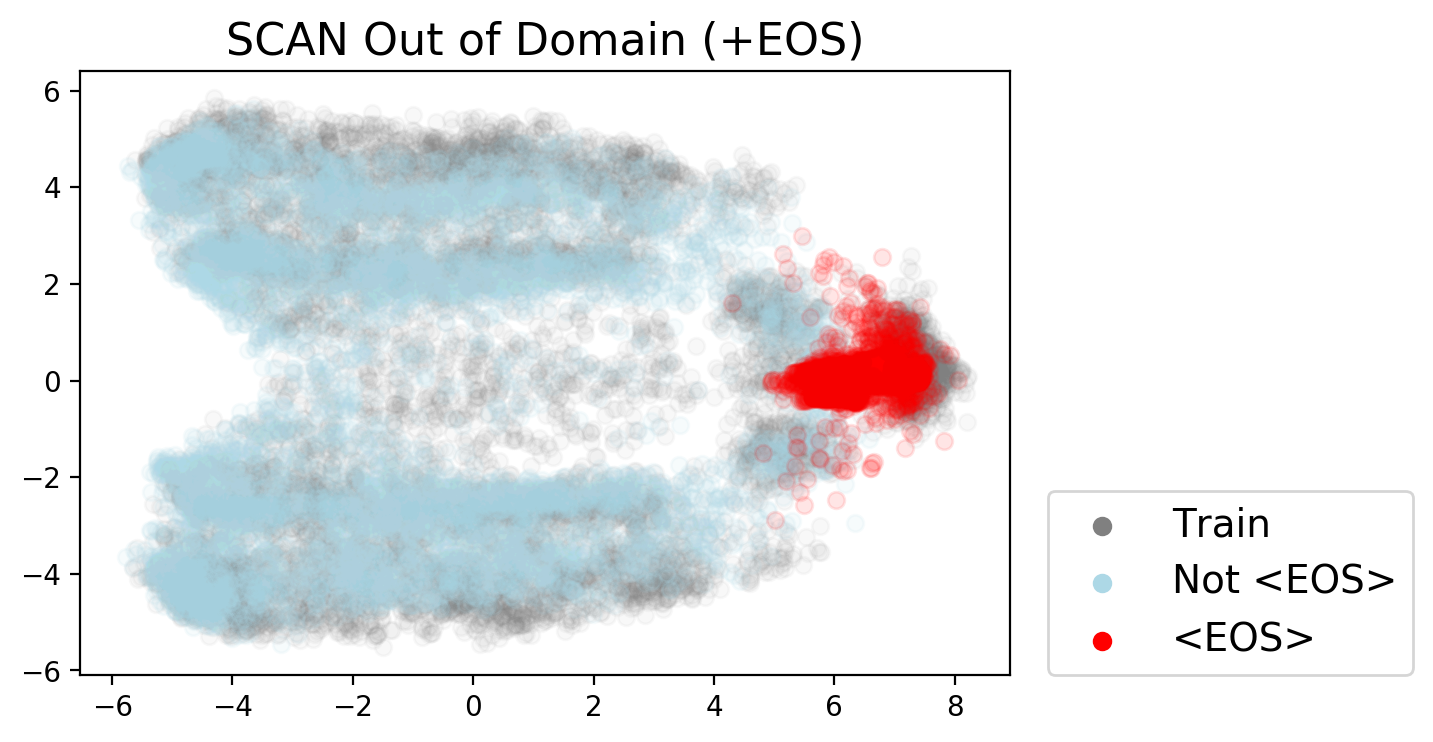}}
    \subcaptionbox{\label{fig:scan-pca-c}}{\includegraphics[width=0.92\linewidth]{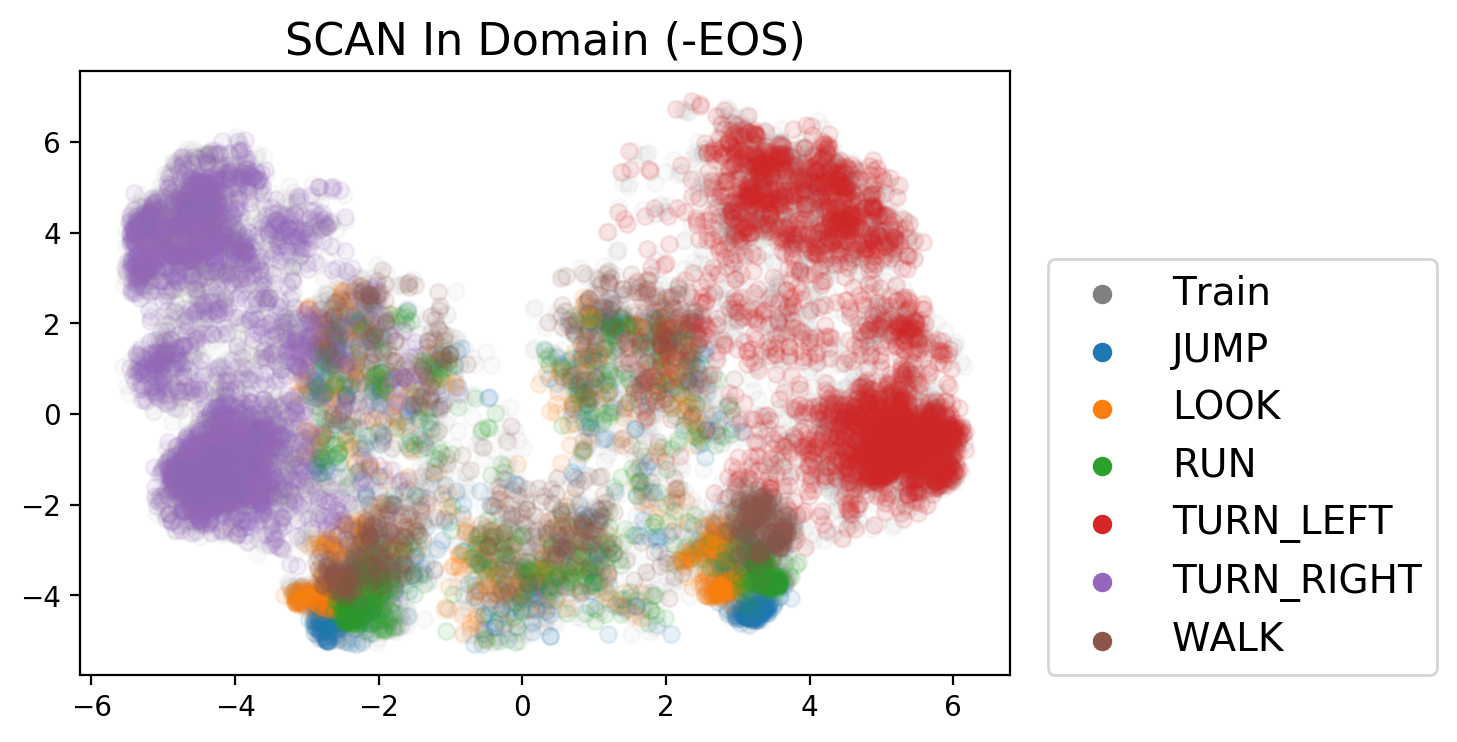}}
    \subcaptionbox{\label{fig:scan-pca-d}}{\includegraphics[width=0.92\linewidth]{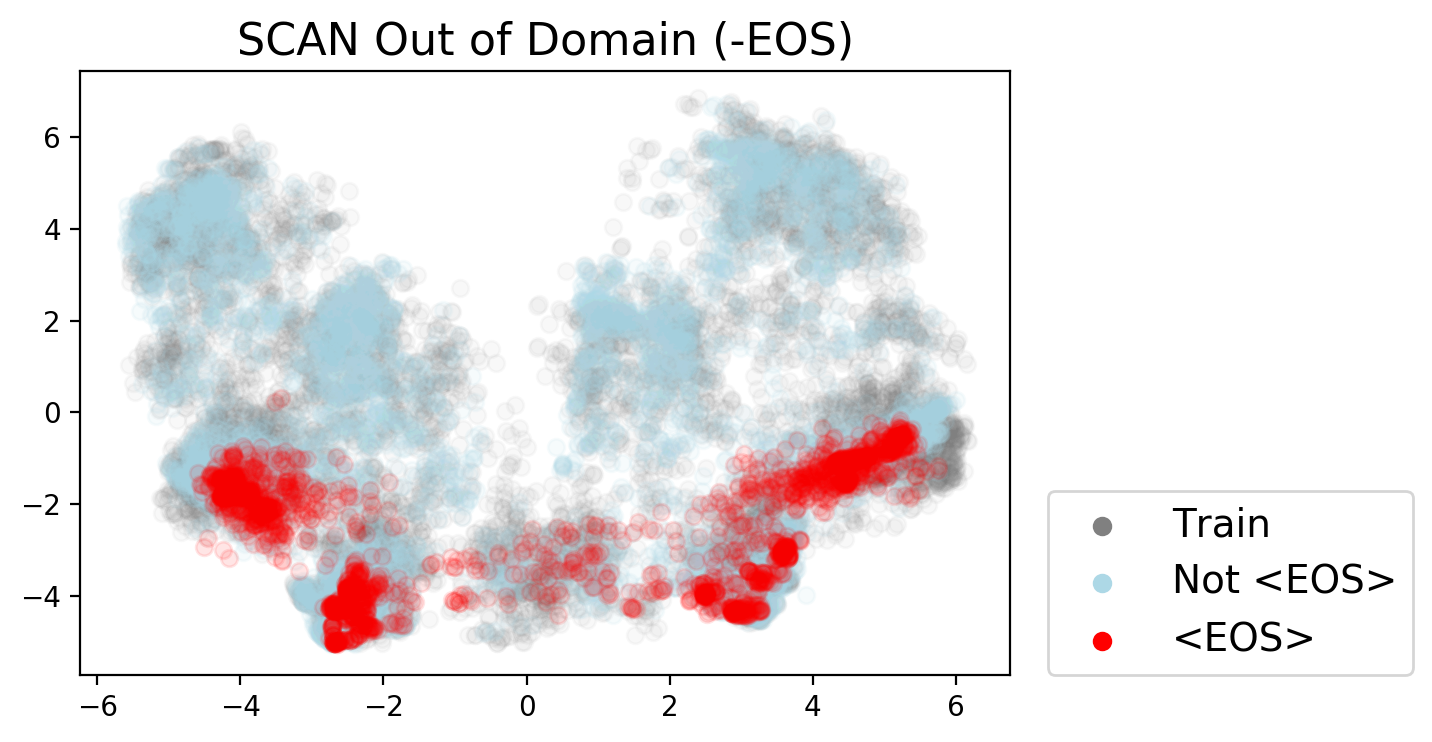}}
    
    \caption{The top two principal components of the hidden states for the \EOSp and \EOSm LSTM models trained on the SCAN 22-token length split. \protect\ref{fig:scan-pca-a} and \protect\ref{fig:scan-pca-c} color hidden states by the identity of the gold output tokens for the \EOSp and \EOSm conditions respectively. \protect\ref{fig:scan-pca-b} and \protect\ref{fig:scan-pca-d} color out-of-domain sequences by whether the \EOSp model puts a plurality of probability mass is put on the \EOS token (\texttt{<EOS>}, red) or on any other token (\texttt{Not <EOS>}, light blue). It is important to note that in both plots, the colors are derived from the \EOSp model as we want to investigate how the \EOSm model performs in the places where the \EOSp model errs.}
    \label{fig:scan-pca} 
\end{figure}

\begin{table*}
    \centering
    \small
    \begin{tabular}{p{3.5cm} c p{8.5cm}}
        \toprule
         Instruction & Model & Predicted Actions \\
         \midrule
         walk around left twice and walk around left thrice & \EOSpOracle & TURN\_LEFT WALK $\ldots$ WALK \color{red}WALK JUMP WALK \color{black}WALK \color{red}JUMP \color{black}WALK \color{red}WALK JUMP \\
        
         & \EOSmOracle & TURN\_LEFT $\ldots$ WALK \color{blue} TURN\_LEFT WALK TURN\_LEFT \color{black}WALK \color{blue}TURN\_LEFT \color{black}WALK \color{blue}TURN\_LEFT WALK \\
         \midrule
         run around left twice and run around right thrice & \EOSpOracle & TURN\_LEFT $\ldots$ TURN\_RIGHT RUN \color{red} RUN \\
         & \EOSmOracle & TURN\_LEFT $\ldots$ TURN\_RIGHT RUN \color{blue} TURN\_RIGHT \\
         \bottomrule
    \end{tabular}
    \caption{Two examples of errors a \EOSp model makes from the 22-token length split: generating irrelevant actions and repeating the last action. Red tokens are incorrect, blue are correct. See the appendix for more examples.
    }
    \label{tab:scan-examples}
\end{table*}

We find that the \EOSmOracle  models consistently outperform \EOSpOracle models across all length splits (Table~\ref{tab:scan-results}). %
We also observe that after including sequences up to length 26, the models have seen enough of the new template to perform with accuracy $\geq 80\%$ on the rest of the long sequences.
However, the question of what the \EOSm model is doing that allows it to succeed remains. %
The \EOSp model fails in the non-oracle setting by predicting that sequences should end before they do, and the \EOSpOracle model fails because once it decodes past its maximum training sequence length, it tends to either repeat the last token produced or emit unrelated tokens.
The \EOSmOracle model succeeds, however, by repeating the last few tokens when necessary, so as to complete the last portion of a \textit{thrice} command for example.
(See Table~\ref{tab:scan-examples} for two such examples).

We compute the top two principal components of the decoder hidden states and plot them as we did for the \dyck{k}{m} models.
While the top two principal components explain a modest amount of the variance (27.78\% for \EOSm and 29.83\% for \EOSp), they are still interpretable (Figure \ref{fig:scan-pca}).
In particular, we can see that the recurrent dynamics break differently in this case compared to the \dyck{2}{4} case.
Here, they break once the \EOSp model puts a plurality of probability mass on the \EOS token (\ref{fig:scan-pca-b}).
Once this happens, the hidden states remain in the cluster of \EOS states (\ref{fig:scan-pca-a}, pink cluster).
The \EOSm condition does not have such an attractor cluster (\ref{fig:scan-pca-d}), so these same hidden states are able to transition between clusters associated with various token identities (\ref{fig:scan-pca-c}).
This freedom likely aids extrapolative performance.

A figure with hidden states colored by length is available in Appendix~\ref{app:plots-scan}.
Both the \EOSp and \EOSm show evidence of tracking token position because knowing how many actions have been taken is an important component of the SCAN task (for example, to follow an instruction like ``turn left twice and walk around right thrice'', a model needs to recall how many times it has turned left in order to correctly continue on to walking around thrice). %

The results we see here shed some light on the length generalization work by \citet{lake2018generalization}. They note that when they prevent a model from predicting \EOS (our \EOSpOracle metric), they achieved $60.2\%$ exact match accuracy on the 22-token length split with a GRU with a small hidden dimension. This number is comparable to the number we find using \EOSmOracle metric. The success of the low-dimensional GRU may very well be related to its failure to implement counters as efficiently as LSTMs \cite{weiss2018practical}---if the model cannot count well, then it may not learn length attractors.

\subsection{Transformer-based models}
All of our experiments so far have used LSTMs, but transformers are the current state of the art in many NLP tasks.
They also can be trained with the standard fixed sinusoidal positional embeddings, which \citet{vaswani2017attention} suggest might help with extrapolation due to their periodicity.
We train transformer \EOSp and \EOSm models with sinusoidal positional embeddings on the SCAN length splits and observe that \EOSp models perform about as poorly as \EOSp models in the LSTM case as well (Table~\ref{tab:scan-results}).
Despite using periodic positional embeddings, \EOSp models are not able to extrapolate well in this setting.

\section{Experiment 3: Machine Translation}
\label{section:mt}
Our final experiment focuses on how well the extrapolation we see in the case of SCAN scales to a human-language task---translation from German to English from the WMT2009 challenge. This data is much more complex than the SCAN task and has many more subtle markers for length, which might act as proxies for the \EOS token, meaning that removing \EOS tokens from the training data has a smaller impact on the models' extrapolative abilities.  We find that there is very little difference between the \EOSp and \EOSm models' performance on out-of-domain lengths compared to SCAN, and while \EOSm perform better in out-of-domain settings more often than \EOSp models, removing the \EOS token does not conclusively help with extrapolation.

\subsection{Methodology}
We use a subset of 500,000 German to English sentences from the WMT2009 challenge Europarl training set. The median English sentence length has 24 tokens; we create three length splits: with $\ell= 10$, $15$, and $25$, giving us approximately 61, 126, and 268 thousand training examples respectively.
We train \EOSp and \EOSm LSTM models with 2 layers for the encoder and decoder with 500 and 1000-dimensional hidden states as well as Transformer models, the result of a hyperparameter search described in Appendix~\ref{app:exp-mt}.
Additionally, the training data in both conditions is modified by removing any final punctuation from the sentences that could serve as explicit markers for length.
Finally, our evaluations are very similar to the ones in our SCAN experiments, but our metric is BLEU rather than exact match accuracy \citep{papineni-etal-2002-bleu}.

We report standard BLEU for the \EOSp condition, and two length oracles, where we prevent the \EOSp models from producing \EOS tokens and force sequences produced by all models to stop decoding at the length that maximizes their BLEU score.
In practice, computing the BLEU score for all sequence lengths is expensive, so we consider only lengths within a window of 7 tokens on either side of the gold target length, which gets us within $\sim$1\% of the true oracle score.
We train our models using OpenNMT \cite{opennmt} and calculate BLEU using the sacreBLEU package for reproducibility \citep{post2018call}.

\subsection{Results}

\begin{table}
    \centering
    \small
    \renewcommand{\tabcolsep}{4pt}
    \begin{tabular}{l   rr   r r   r r}
         \toprule
         $\ell$ & \multicolumn{2}{c}{10} & \multicolumn{2}{c}{15} & \multicolumn{2}{c}{25} \\
         \midrule
         LSTM  & ID & OOD & ID & OOD & ID & OOD \\
         \it\EOSp & \it25.25 & \it1.75& \it25.27 & \it7.24 & \it28.14 & \it16.19 \\
         \scriptsize \EOSpOracle &26.42 & 4.64 & \bf 26.43 & 11.75 & \bf 29.01 & \bf 20.34 \\
         \scriptsize\EOSmOracle &\bf26.59 & \bf5.14 & 25.84 & \bf 12.53 & 28.70 & 20.12 \\
         $\Delta$ & 0.17 & 0.5 & -0.59 & 0.78 & -0.31 & -0.22 \\
         \midrule
         \midrule
         Transformer&  &  &  &  \\
         \it\EOSp & \it24.91 & \it1.27 & \it25.67 & \it5.16 & \it 28.75 & \it 13.32 \\
         \scriptsize \EOSpOracle & 26.15 & \bf4.87 & 26.33 & 10.37 & \bf29.29 & 17.14 \\
         \scriptsize \EOSmOracle & \bf26.73 & 4.65 &\bf26.81 & \bf11.65  & 29.13 & \bf17.39 \\
         $\Delta$ & 0.58 & -0.22 & 0.51 & 1.32 & 0.16 & 0.25 \\
        \bottomrule
    \end{tabular}
    \caption{German-to-English translation BLEU scores. ID is on held-out data with the same length as the training data and OOD is data longer than that seen at training.}
    \label{tab:mt}
\end{table}

We observe a slight increase in extrapolative performance for \EOSm models over \EOSp models for LSTMs in the 15 and 10-token length splits, and transformers in the 15 and 25-token length splits, but have no consistent takeaways (Table~\ref{tab:mt}).

We also report in-domain BLEU scores.
There is some variation between these, but mostly less than between the out-of-domain scores, which may suggest that the difference of extrapolative performance in those models is meaningful.
Additionally, we do not report plots of the top two principal components for these models because they only explain $3\%$ of the variance and are not visually interesting.

We speculate that we do not see the \EOSm models consistently outperforming the \EOSp ones because are likely more subtle indicators of length that models in both conditions pick up on, rendering the presence of \EOS tokens less relevant.
Further analysis should look to mitigate these length cues as well as investigate additional length splits.

\section{Discussion}
For most purposes of generative modeling, it is necessary to know when to end the generative process in order to use a model; put another way, a \EOSm model is not usable by itself.
The immediate engineering question is whether there exists a way to learn a distribution over when to end the generative process that does not have the same negative effects as the \EOS decision.

In our \dyck{k}{m} experiments, we observed that even in \EOSm models, there exists a linear separator in hidden state space between points where the generative process is and isn't allowed to terminate.
We explore training probe to find this linear separator, and are able to predict where sequences end with high accuracy (Table~\ref{tab:dyck-probe}).
In our other experiments, we did not see such a simple possible solution, but can speculate as to what it would require.
In particular, the length manifold and length attractor behaviors seem to indicate that length extrapolation fails because the conditions for stopping are tracked in these models more or less in terms of absolute linear position.
As such, it is possible that a successful parameterization may make use of an implicit checklist model \cite{kiddon2016globally}, that checks off which parts of the input have been accounted for in the output.

Can we use the same neural architectures, in the \EOSp setting, while achieving better length extrapolation?
Our PCA experiments seem to indicate that \EOSm models' length tracking is (at least) linearly decodable.
This suggests that training models with an adversarial loss %
against predicting the position of the token may encourage \EOSp models to track length in a way that allows for extrapolation. %

\section{Conclusion}
In this work, we studied a decision often overlooked in NLP: modeling the probability of ending the generative process through a special token in a neural decoder's output vocabulary.
We trained neural models to predict this special \EOS token and studied how this objective affected their behavior and representations across three diverse tasks.
Our quantitative evaluations took place in an oracle setting in which we forced all models to generate until the optimal sequence length at test time.
Under this setting, we consistently found that networks trained to predict the \EOS token (\EOSp) had a worse length-extrapolative ability than those not trained to (\EOSm).
Examining the hidden states of \EOSp and \EOSm networks, we observed that \EOSp hidden states exhibited \textit{length manifolds} and \textit{length attractor} behaviors that inhibit extrapolation, and which otherwise identical \EOSm networks do not exhibit.
Thus, we argue that training to predict the \EOS token causes current models to track generated sequence length in their hidden states in a manner that does not extrapolate out-of-domain. %
When \EOS tokens were first introduced to NLP ensure probabilistic discrete sequence models maintained well-formed distributions over strings, these models, with a small sets of hidden states, did not readily pick up on these length correlations.
However, when this NLP technique was ported to more expressive neural networks, it hid potentially useful length-extrapolative inductive biases.
We see the evidence presented here as a call to explore alternative ways to parameterize the end of the generative process.

\section*{Acknowledgments}
The authors thank Nelson Liu and Atticus Geiger for comments on early drafts, and to our reviewers whose helpful comments improved the clarity of this work. 
JH was supported by an NSF Graduate Research Fellowship under grant number DGE-1656518.

\bibliographystyle{acl_natbib}
\bibliography{anthology,emnlp2020}

\newpage
\appendix

\section{Experiment Details}
\label{app:exp}

\subsection{\dyck{k}{m}}
\label{app:exp-dyck}
The language \dyck{k}{m}, for $k,m\in \mathbb{Z}^{+}$, is a set of strings over the vocabulary consisting of $k$ open brackets, $k$ close brackets, and the \EOS token.
To generate a dataset for \dyck{k}{m}, we must thus define a distribution over the language, and sample from it.
Defining this distribution characterizes the statistical properties of the language -- how deeply nested is it on average, how long are the strings, etc.

At a high level, we wanted our samples from \dyck{k}{m} to be difficult -- that is, to require long-term memory, and to traverse from empty memory to its maximum nesting depth ($m$) and back multiple times in the course of a single sample, to preclude simple heuristics like remembering the first few open brackets to close the last few.
For further discussion of evaluation of models on Dyck-$k$ and similar formal languages, see \citet{suzgun2019evaluating}.

The distribution we define is as follows.
If the string is balanced, we end the string with probability $1/2$ and open any of the $k$ brackets with probability $1/2$.
If the string has more than $0$ unclosed open parentheses but fewer than $m$ (the bound), we open a bracket with probability $1/2$ and close the most recently opened bracket with probability $1/2$; if we open a bracket, we choose uniformly at random from the $k$.
If the string has $m$ unclosed open brackets, then it has reached its nesting bound, and we close the most recently opened bracket with probability $1$.

Sampling directly from this distribution would lead to exponentially short sequences, which would break our desideratum of difficulty.
So, we truncate the distribution by the length of strings, ensuring a minimum (and a maximum) length at training time.

We define the length truncation as follows.
Consider that the number of unclosed open brackets at timestep $t$ is some number between $0$ and $m$.
At any timestep, we move from $s_i$ to $s_{i+1}$ with probability 1/2 (opening a bracket), or from $s_i$ to $s_{i-1}$ with probability 1/2 (closing a bracket.)
Let these states be $s_0$ to $s_m$.
What we'd like to see is that we eventually move from state $s_0$ to $s_m$ and back to $s_0$ multiple times, making sure that the network cannot use positional heuristics to remember the first few brackets in order to close the last few.
We ensure this in expectation by making sure that the minimum truncation length of samples is defined so that in expectation, sampled strings traverse the Markov chain from $s_0$ to $s_m$ and back to $s_0$ at least three times.

In practice, we set the minimum length of training samples to $6m(m-2)+40$, and the maximum length to $7m(m-2)+60$.
Note the quadratic dependence on $m$; this is because of the expected hitting time of $s_0$ to $s_{m+1}$ of the $m+1$-state Markov chain wherein there is a $1/2$ probability of moving in each direction.
We add constant factors to ensure there are no extremely short sequences for small $m$; the exact truncation lengths are not as important as the general scaling.

Additionally, as $m$ increases, the number of training examples required for consistently high in-domain performance also increases. We use $10^{\frac{m}{2}+2}$ training examples for each of the values of $m$. For all values of $m$, we use $10000$ test samples.

Our models are single-layer LSTMs with $5m$ hidden states. We a batch size of $2^{\frac{m}{2} + 2}$, and use the Adam optimizer with a learning rate of $0.01$ \citep{kingma2014adam}.

\subsection{SCAN}
The SCAN model we train is the model that achieves the best accuracy in-domain in the experiments of \citet{lake2018generalization}.
This is an LSTM sequence-to-sequence model with 2 layers of 200-dimensional hidden units and no dropout or attention.
Optimization was done with the Adam optimizer \citep{kingma2014adam}, learning rate of $0.001$.
We used OpenNMT to train the transformer models \cite{opennmt}. These models have $2$ layers, $128$-dimensional hidden states, $8$ attention-heads, and $1024$-dimensional feedfoward layers. All other hyperparameters were the defaults suggested by OpenNMT for training transformer sequence-to-sequence models.

\subsection{Machine Translation}
\label{app:exp-mt}
The German-English MT models we train were also trained with Open NMT. We trained LSTM models with attention and $1000$-dimensional and $500$-dimensional hidden states and two layers. All other parameters were specified by the OpenNMT defaults as of \texttt{OpenNMT-py} version 1.1.1.
For the $25$-token length split we report results with $1000$-dimensional hidden states, and with $500$-dimensional hidden states for the $10$-length split.
For the $15$-token length split the \EOSp model used $1000$-dimensional hidden states while the \EOSpOracle and \EOSmOracle models used $500$-dimensional hidden states.
This difference is a result of running hyperparameter optimization and finding that \EOSp baseline performed better with a larger hidden state size.
Our transformer models were trained using the hyperparameters suggested by OpenNMT for training transformer sequence-to-sequence models. This includes 6 layers for the decoder and decoder, $512$-dimensional hidden states and $2048$-dimensional feed-forward layers.

\section{Additional Plots and Results for Dyck-2,m Languages}
\label{app:plots-dyck}

\begin{figure}[!ht]
    \centering
    \includegraphics[width=0.8\linewidth]{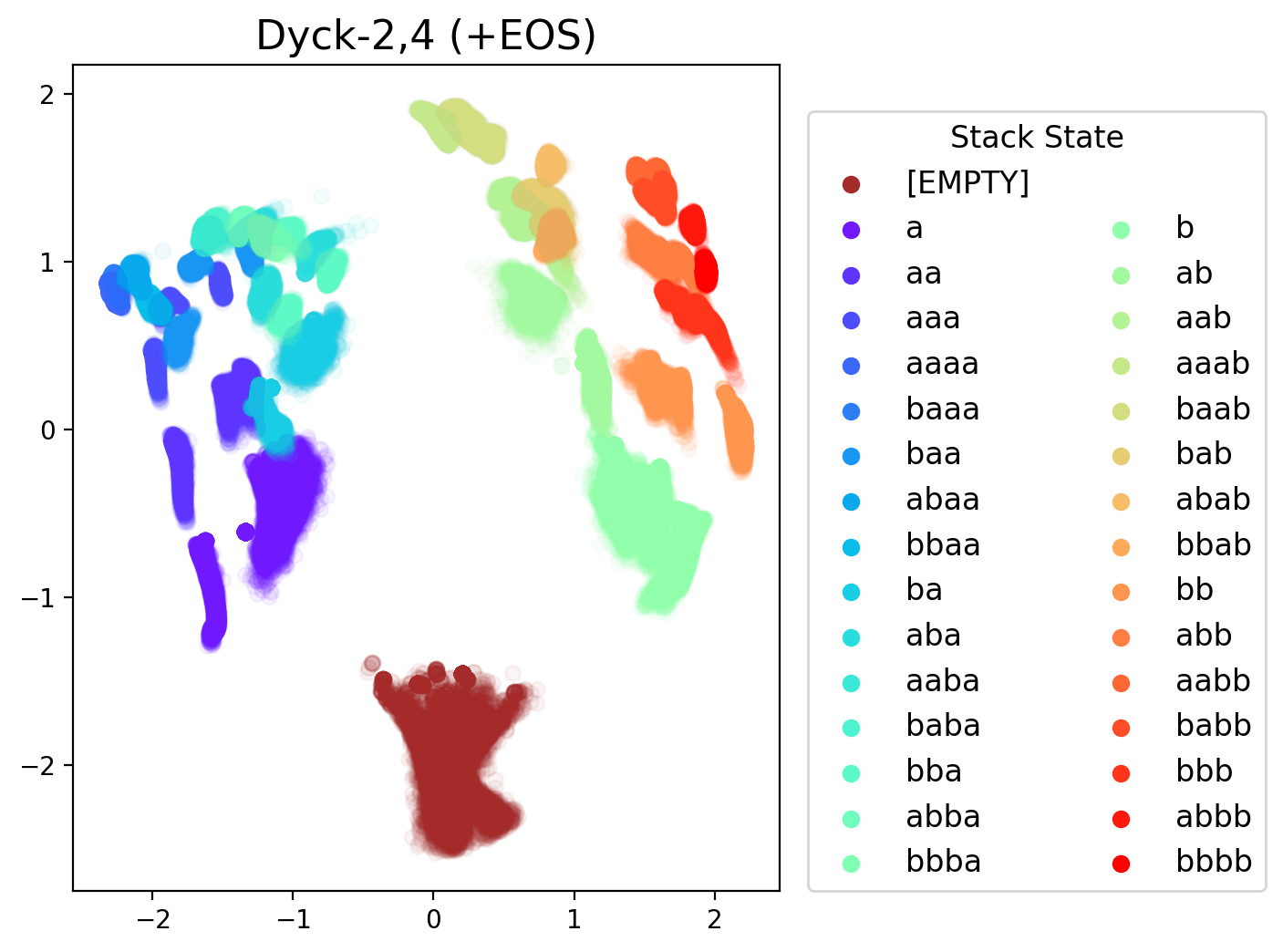}
    \includegraphics[width=0.8\linewidth]{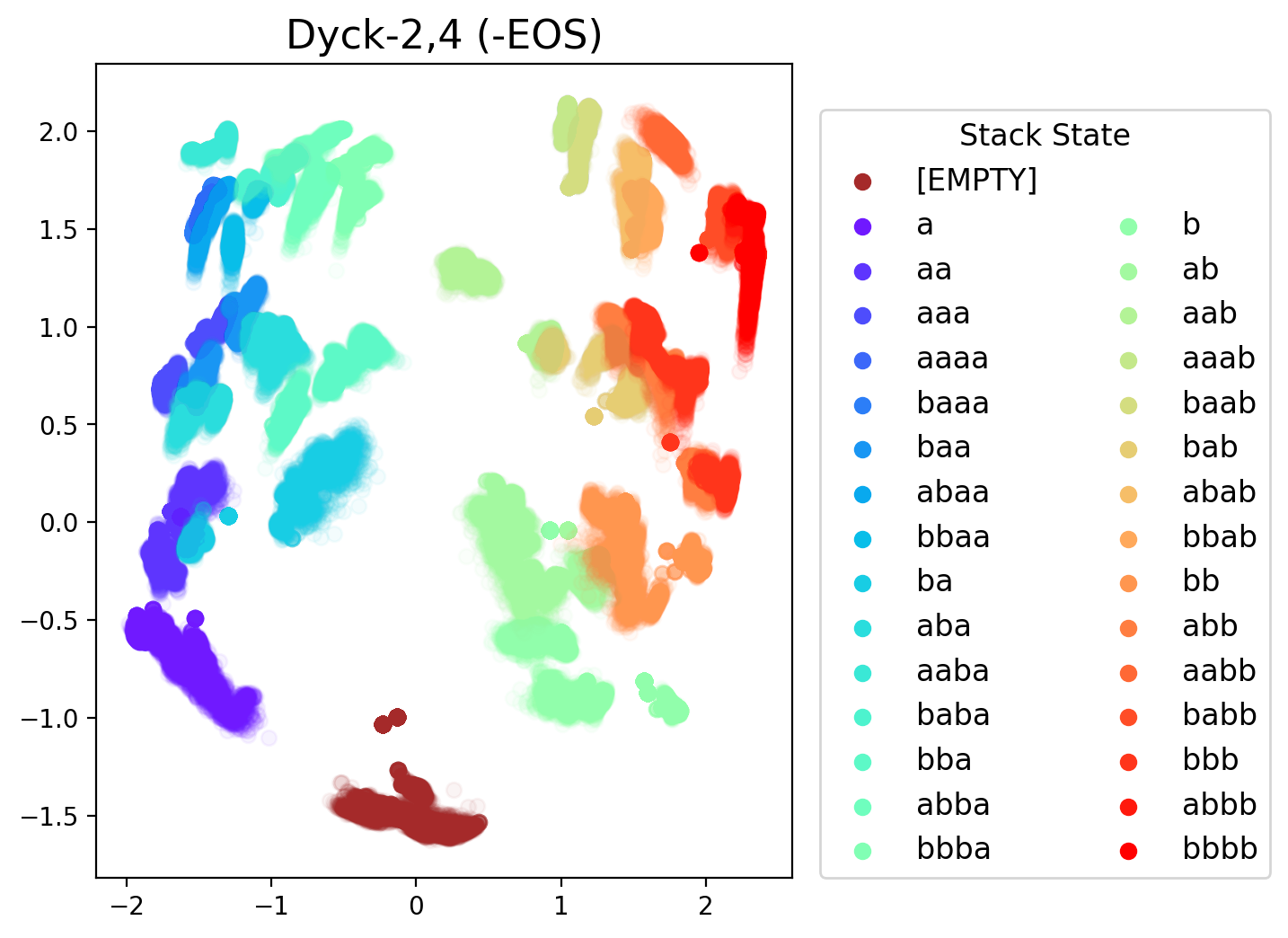}
    \caption{Above we can see a plot of the top two principal components of hidden states from the \EOSp and \EOSm LSTMs trained on they Dyck-2,4. These hidden states come from the training samples. The hidden states are colored by stack state---at that point in the sequence, what brackets must be closed and in what order. Because this is Dyck-2,4 there are 2 types of brackets which we can call ``a'' and ``b''. An ``a'' in stack state in the legend represents that there is a bracket of type ``a'' on the stack, and the same for ``b''. The top of the stack is the right-most character. We can see that the the first principal component roughly represents whether an ``a'' or ``b'' is at the top of the stack.}
    \label{fig:plots-dyck24}
\end{figure}

\begin{table}
    \centering
    \begin{tabular}{c c c c c}
        \toprule
         m & \multicolumn{2}{c}{\EOSp} & \multicolumn{2}{c}{\EOSm}  \\
         \midrule
           & ID & OOD & ID & OOD \\
         4 & 2.39 & 7.09 & 2.38 & \bf2.93\\
         6 & 2.52 & 5.14 & 2.52 & \bf2.60 \\
         8 & 2.60 & 5.24 & 2.59 & \bf2.68 \\
         \bottomrule
    \end{tabular}
    \caption{Perplexities of the \EOSp and \EOSm models trained on Dyck-2,m languages on in-domain (ID) and out-of-domain (OOD) strings. The perplexities support the bracket closing results---the perplexities are very similar in-domain and out-of-domain for \EOSm model but differ substantially for the \EOSp model. Reported value is the median of five runs, lower is better.}
    \label{tab:dyck-ppl}
\end{table}

Our \EOSm models were not able to achieve perfect accuracy on the the bracket-closing task on longer sequences. We suspect that this failure is due to the \EOSm models picking up on other signals for sequence length that were present in our \dyck{2}{m} samples. Namely, all of our samples ended in an empty bracket state, so the final non-\EOS token was always a closing bracket. To address this issue, we explored removing a random number of additional tokens from the end of the \EOSm models' training data (\EOSm+Random Cutoff). We found that doing this allowed us to perfectly extrapolate to sequences 10x longer (Table~\ref{tab:dycklengths-randomcutoff}).

\begin{table}
    \centering
    \small
    \begin{tabular}{c c c c}
    \toprule
         $m$ & \EOSp & \EOSm & \thead{\EOSm\\ + Random Cutoff}\\
         \midrule
         4  & $0.60$ & $0.86$ & $\mathbf{1.0}$ \\
         6 & $0.68$ & $0.98$ & $\mathbf{1.0}$ \\
         8 & $0.68$ & $0.96$ & $\mathbf{1.0}$ \\
    \bottomrule
    \end{tabular}
    \caption{\dyck{2}{m} bracket closing metric results, median of 5 independent training runs.}
    \label{tab:dycklengths-randomcutoff}
\end{table}

\section{Additional Plots and Samples for SCAN Task}
\label{app:plots-scan}

\begin{figure}[!ht]
    \centering
    \includegraphics[width=0.8\linewidth]{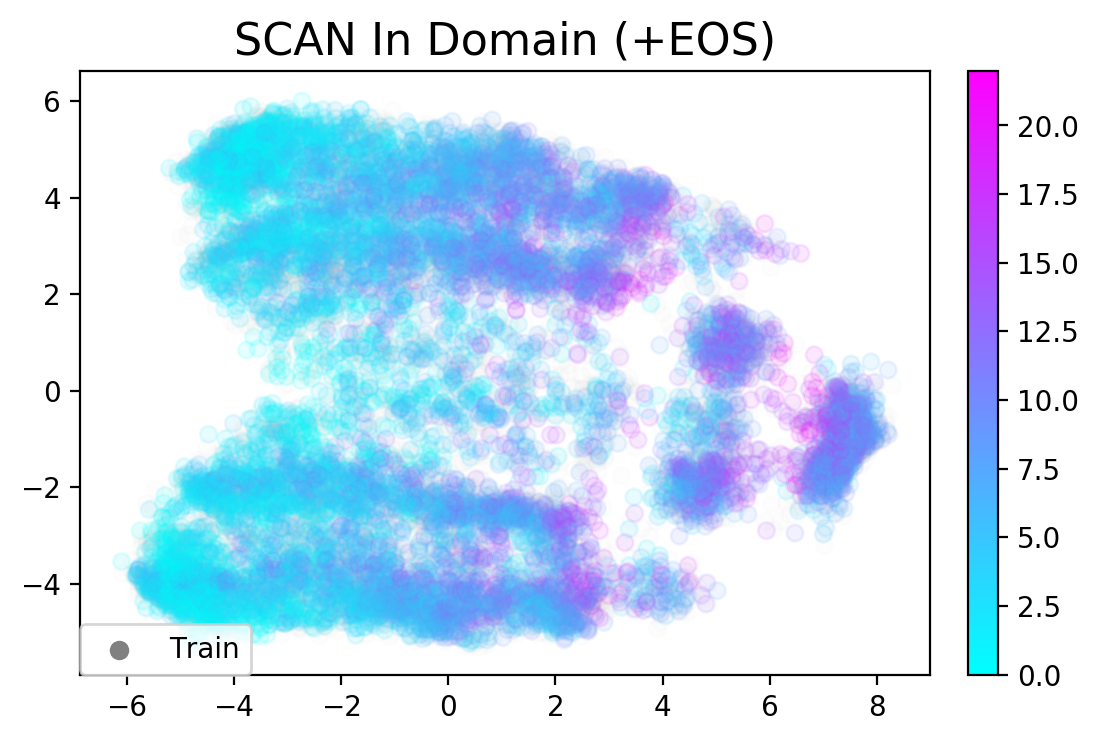}
    \includegraphics[width=0.8\linewidth]{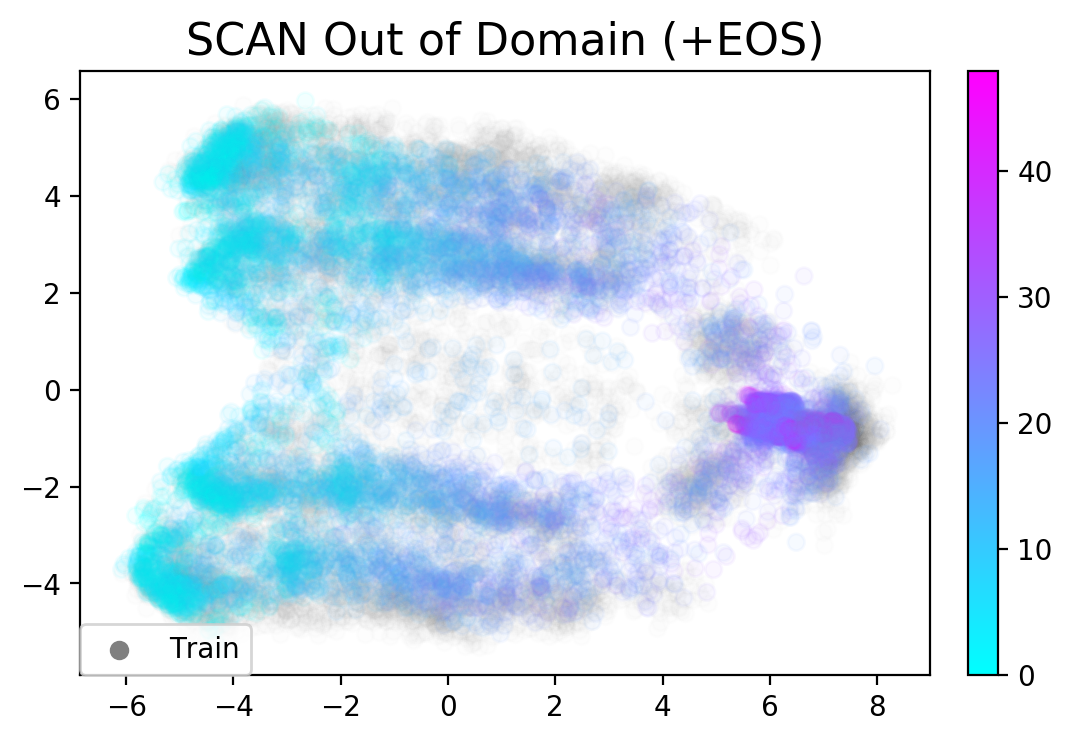}
    \includegraphics[width=0.8\linewidth]{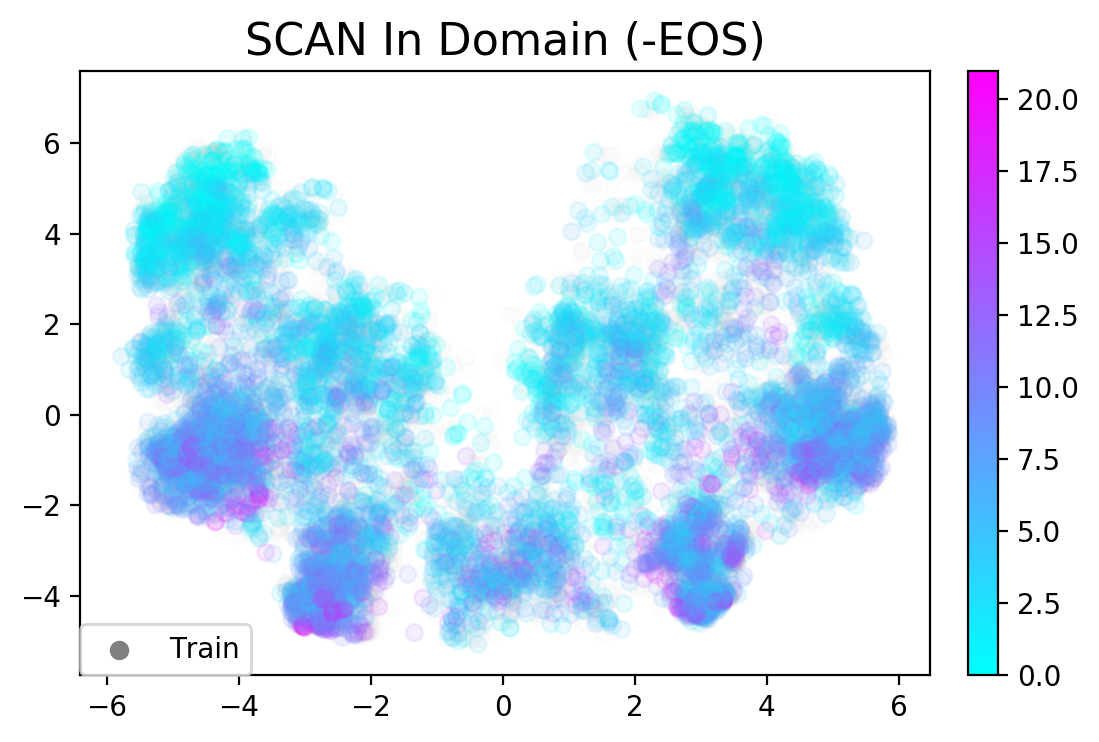}
    \includegraphics[width=0.8\linewidth]{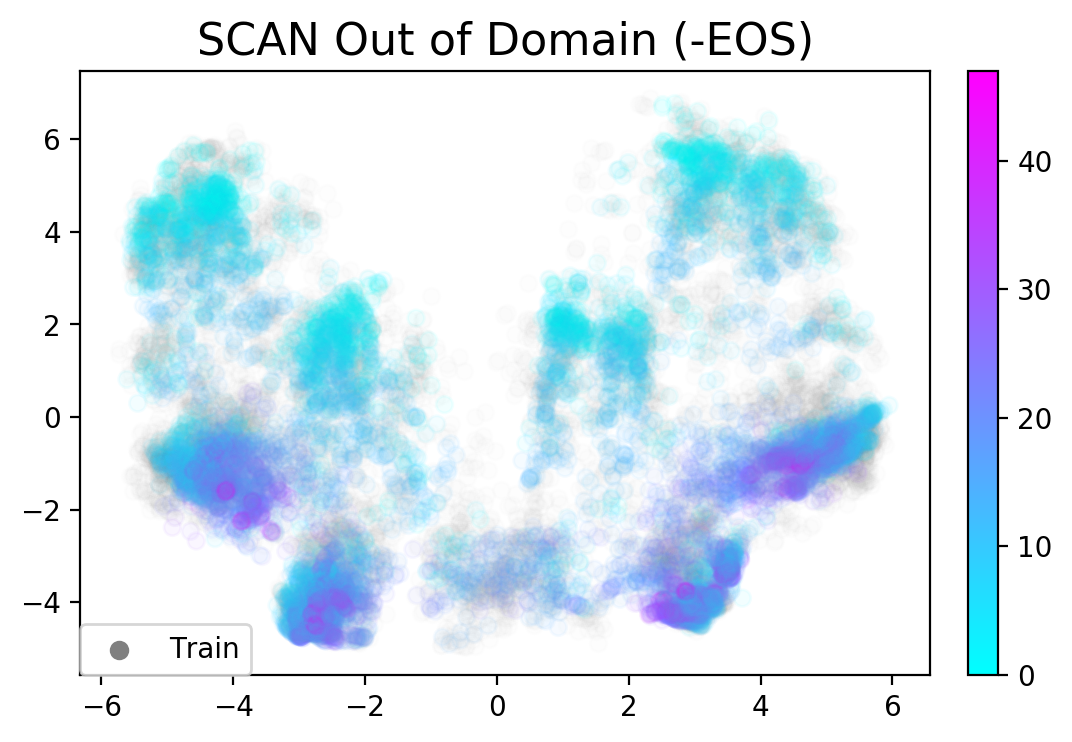}
    \caption{The top two principal components of hidden states from \EOSp and \EOSm LSTMs trained on the SCAN $22$-token length split. Hidden states are colored by their position in the sequence. We can see that both models track length somewhat as it is required, however the \EOSp model has an attractor cluster where all points end up while the \EOSm model does not have such a cluster.}
    \label{fig:scan-lens}
\end{figure}

\begin{figure*}
    \captionsetup[subfloat]{ labelformat=empty}
    \centering
    \subfloat[][Input: run twice after jump right\\Output: \texttt{TURN\_RIGHT JUMP RUN RUN} \label{fig:scan-ex-a}]{\includegraphics[width=0.23\linewidth]{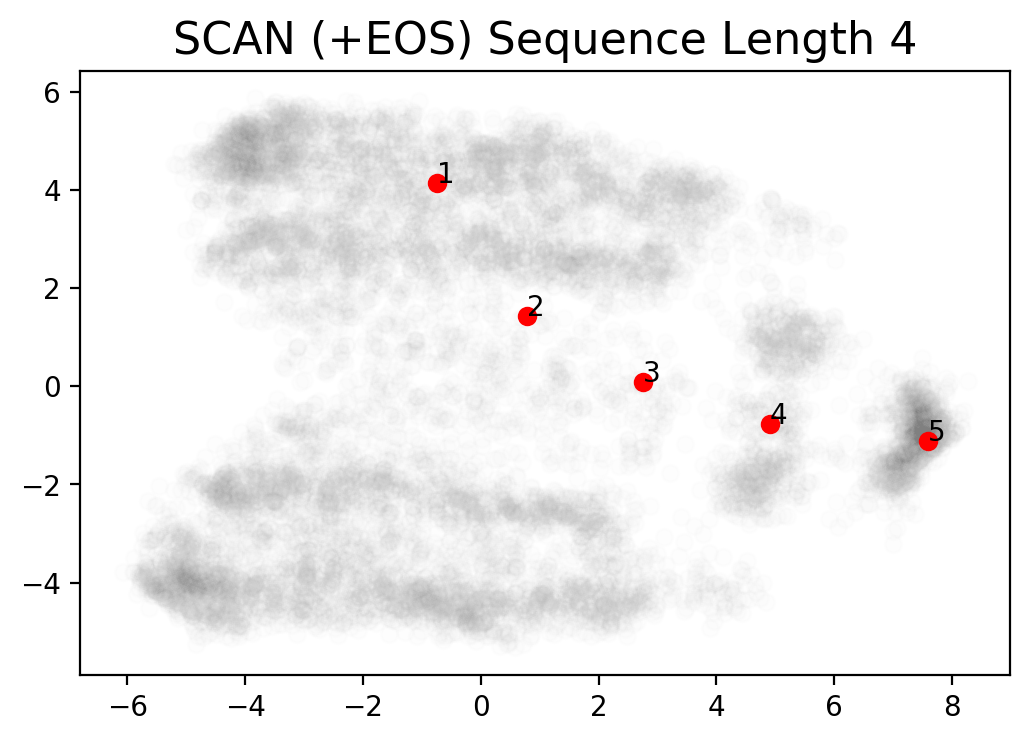}
    \includegraphics[width=0.23\linewidth]{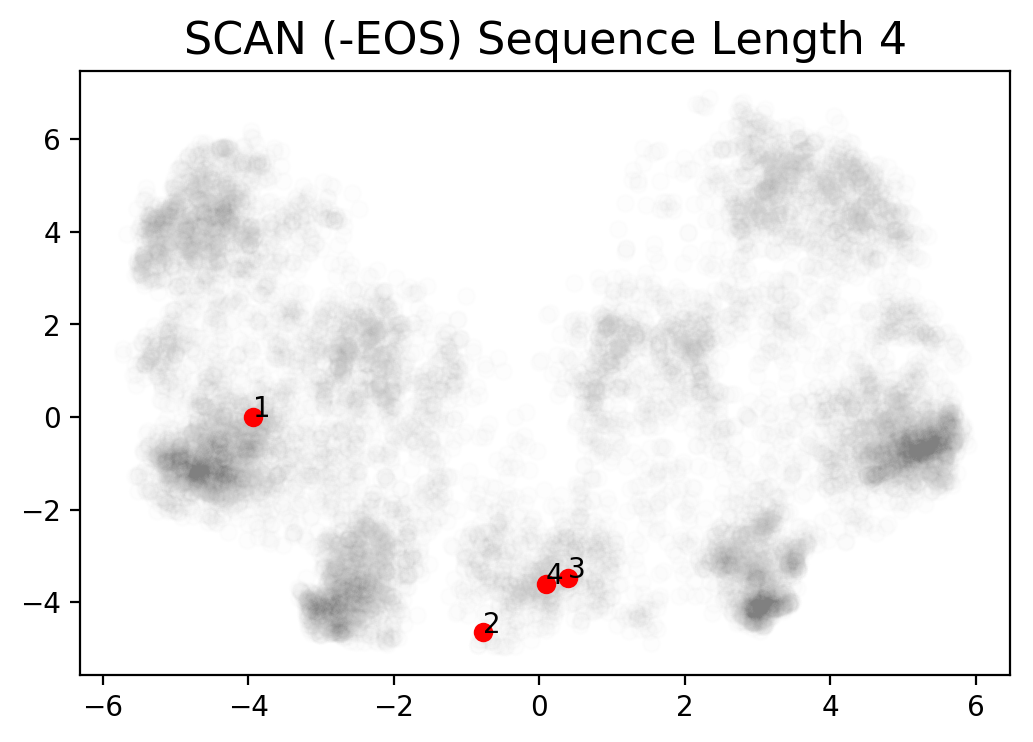}}\;
    \subfloat[][Input: run around right twice \\Output: \texttt{TURN\_RIGHT RUN TURN\_RIGHT RUN TURN\_RIGHT RUN TURN\_RIGHT RUN TURN\_RIGHT RUN TURN\_RIGHT RUN TURN\_RIGHT RUN TURN\_RIGHT RUN}\label{fig:scan-ex-b}]{\includegraphics[width=0.23\linewidth]{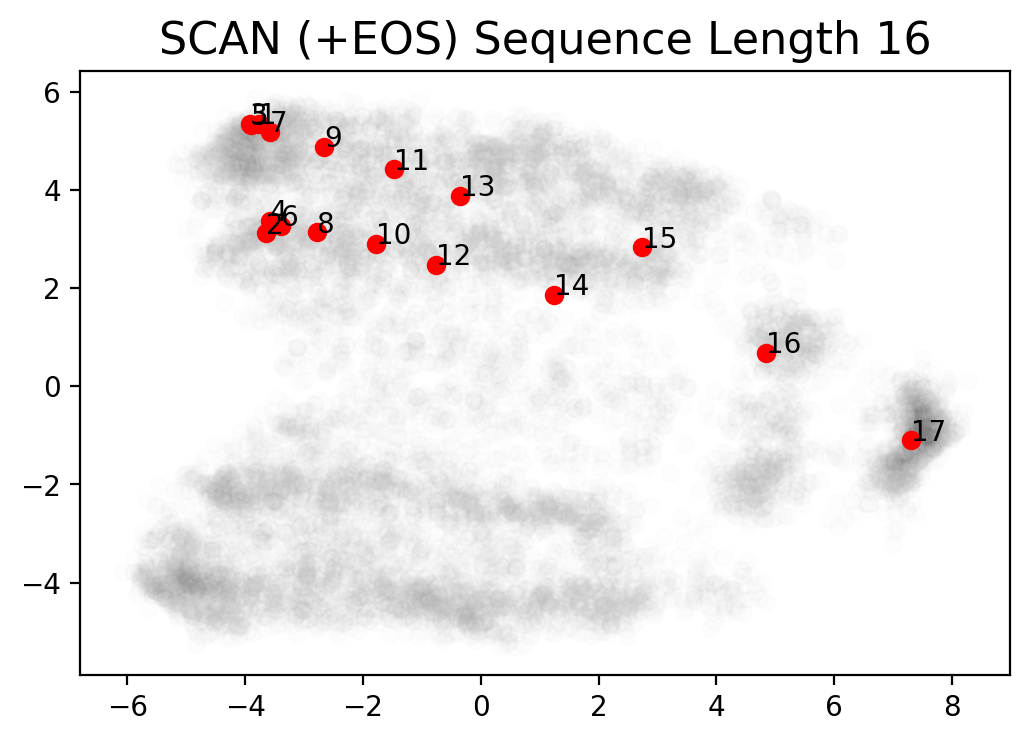}
    \includegraphics[width=0.23\linewidth]{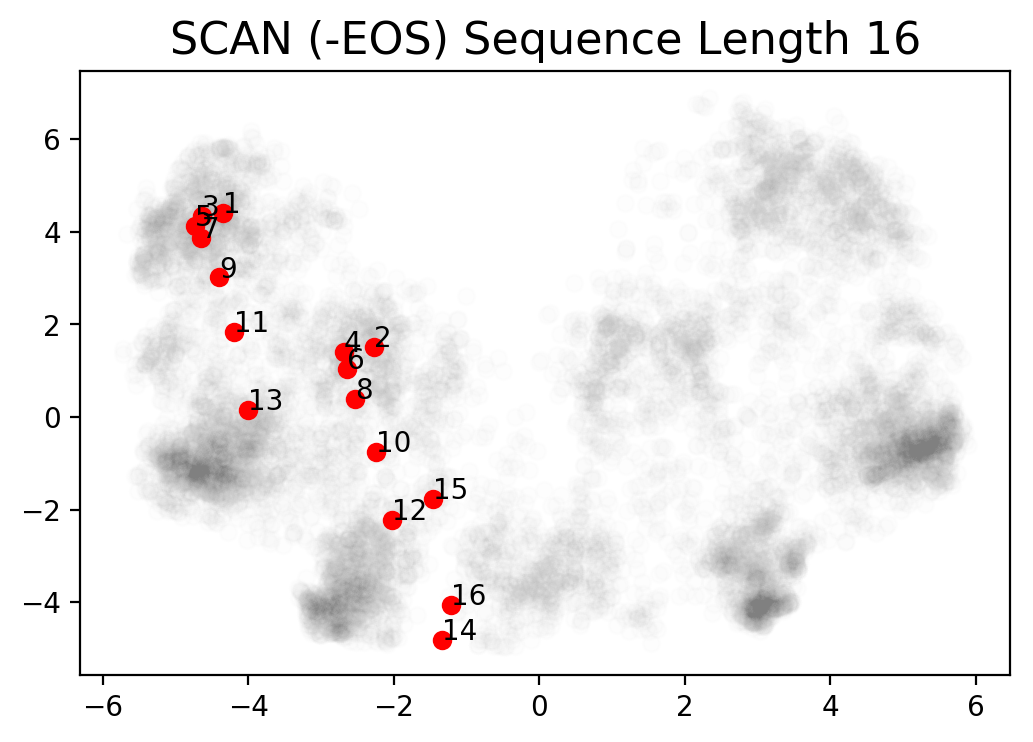}}
    
    \subfloat[][Input: turn around left twice after run around left twice \\Output: \texttt{TURN\_LEFT RUN TURN\_LEFT RUN TURN\_LEFT RUN TURN\_LEFT RUN TURN\_LEFT RUN TURN\_LEFT RUN TURN\_LEFT RUN TURN\_LEFT RUN TURN\_LEFT TURN\_LEFT TURN\_LEFT TURN\_LEFT TURN\_LEFT TURN\_LEFT TURN\_LEFT TURN\_LEFT}\label{fig:scan-ex-c}]{\includegraphics[width=0.23\linewidth]{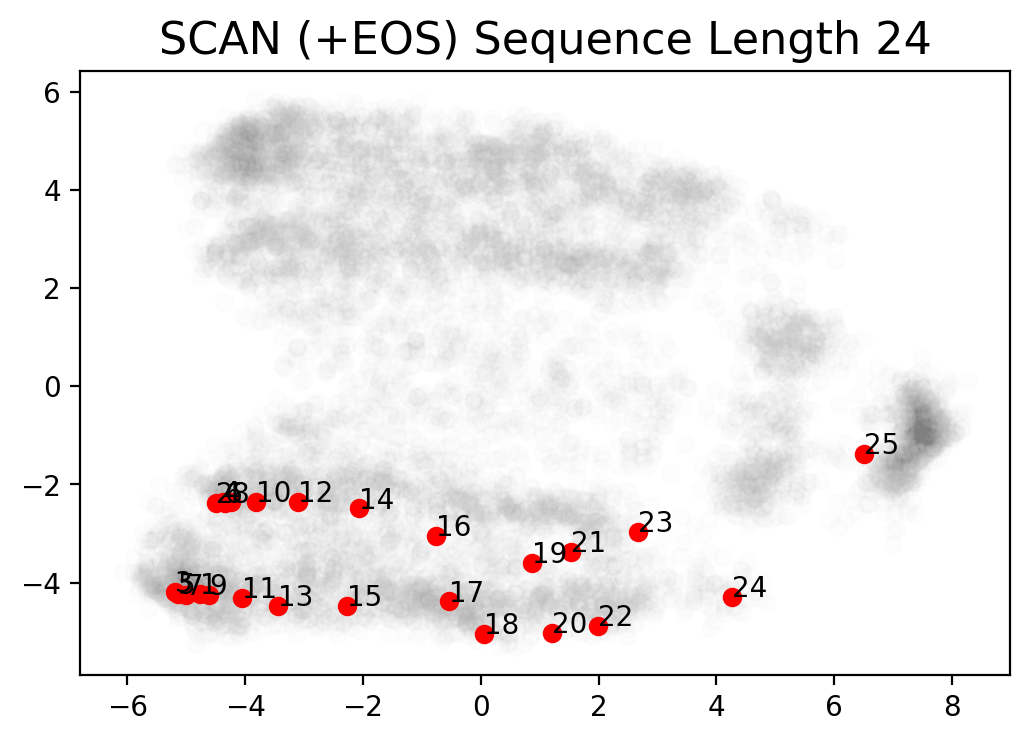}
    \includegraphics[width=0.23\linewidth]{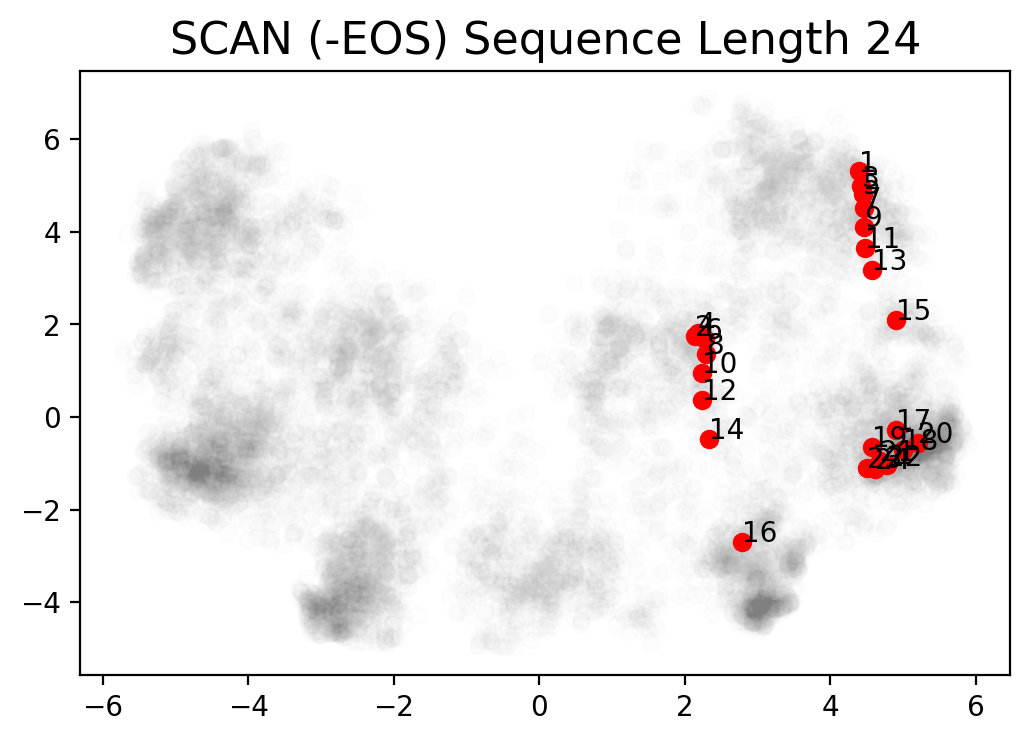}}\;
    \subfloat[][Input: walk opposite left thrice and run around right thrice \\Output: \texttt{TURN\_LEFT TURN\_LEFT WALK TURN\_LEFT TURN\_LEFT WALK TURN\_LEFT TURN\_LEFT WALK TURN\_RIGHT RUN TURN\_RIGHT RUN TURN\_RIGHT RUN TURN\_RIGHT RUN TURN\_RIGHT RUN TURN\_RIGHT RUN TURN\_RIGHT RUN TURN\_RIGHT RUN TURN\_RIGHT RUN TURN\_RIGHT RUN TURN\_RIGHT RUN TURN\_RIGHT RUN}\label{fig:scan-ex-d}]{\includegraphics[width=0.23\linewidth]{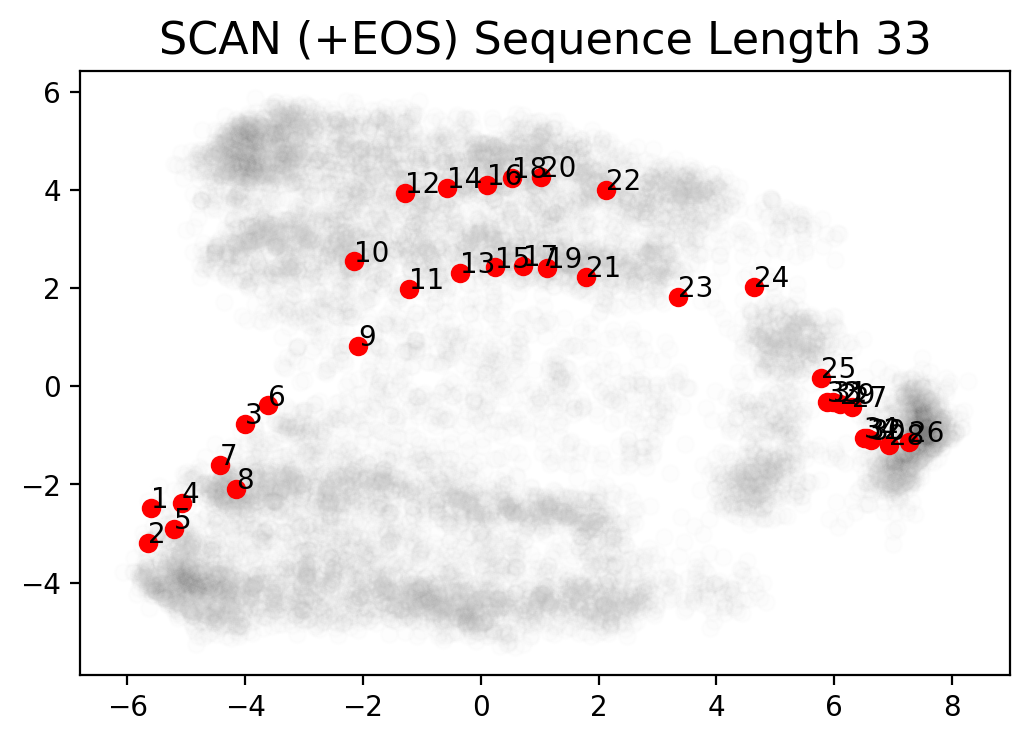}
    \includegraphics[width=0.23\linewidth]{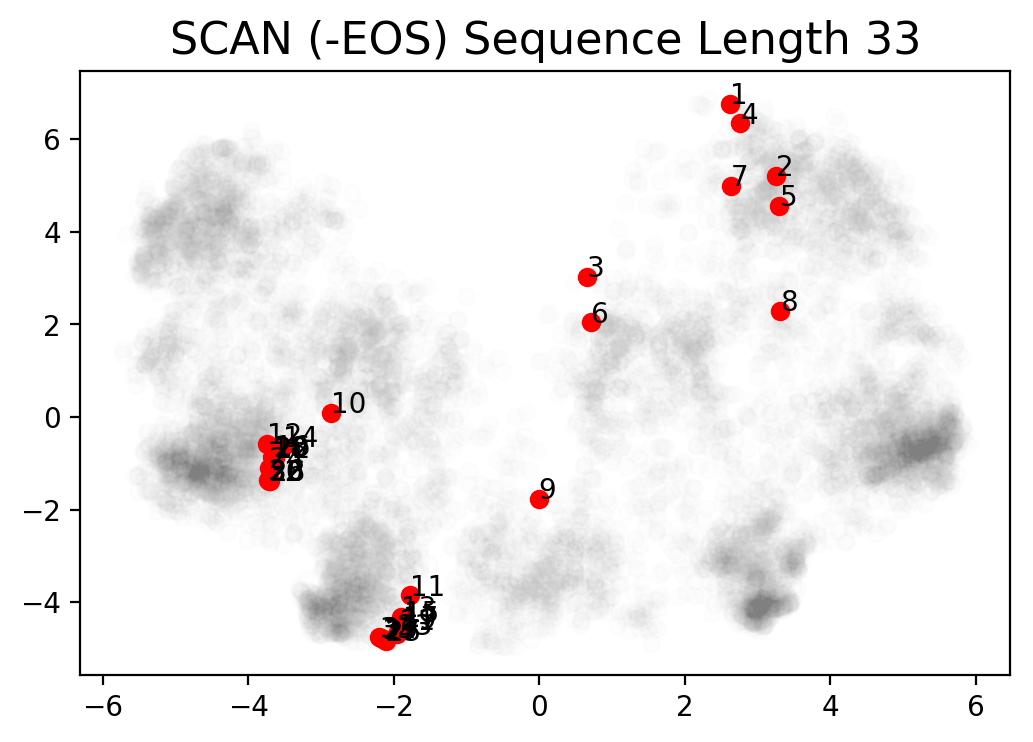}}
    
    \subfloat[][Input: jump around right thrice and walk around right thrice \\Output:\texttt{TURN\_RIGHT JUMP TURN\_RIGHT JUMP TURN\_RIGHT JUMP TURN\_RIGHT JUMP TURN\_RIGHT JUMP TURN\_RIGHT JUMP TURN\_RIGHT JUMP TURN\_RIGHT JUMP TURN\_RIGHT JUMP TURN\_RIGHT JUMP TURN\_RIGHT JUMP TURN\_RIGHT JUMP TURN\_RIGHT WALK TURN\_RIGHT WALK TURN\_RIGHT WALK TURN\_RIGHT WALK TURN\_RIGHT WALK TURN\_RIGHT WALK TURN\_RIGHT WALK TURN\_RIGHT WALK TURN\_RIGHT WALK TURN\_RIGHT WALK TURN\_RIGHT WALK TURN\_RIGHT WALK}\label{fig:scan-ex-e}]{\includegraphics[width=0.23\linewidth]{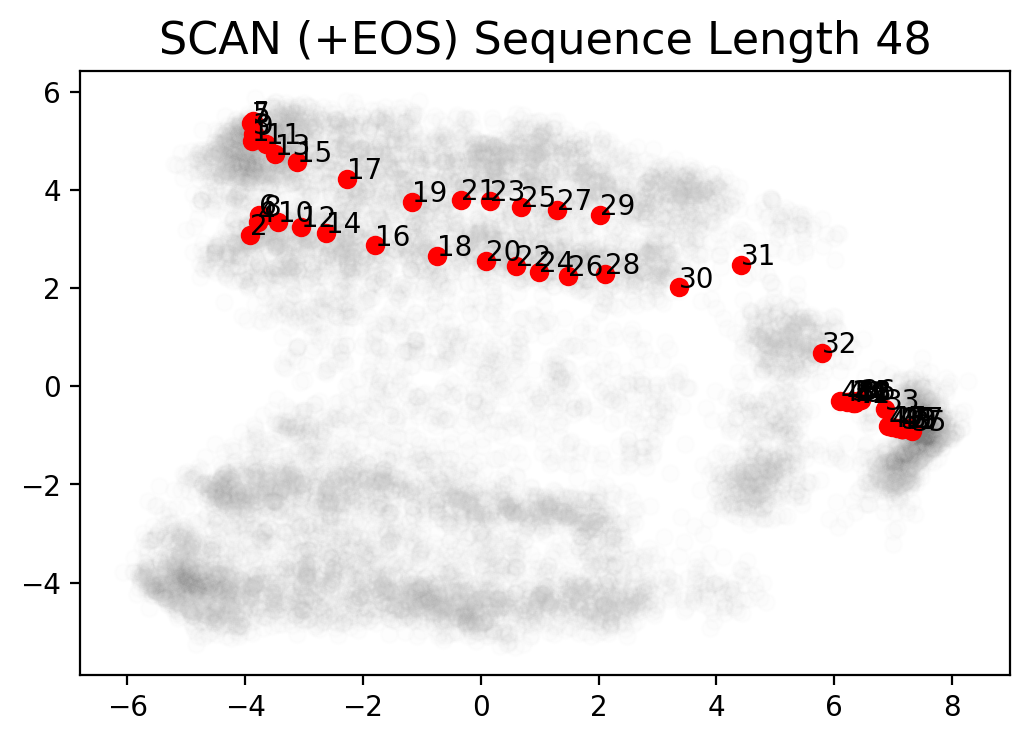}
    \includegraphics[width=0.23\linewidth]{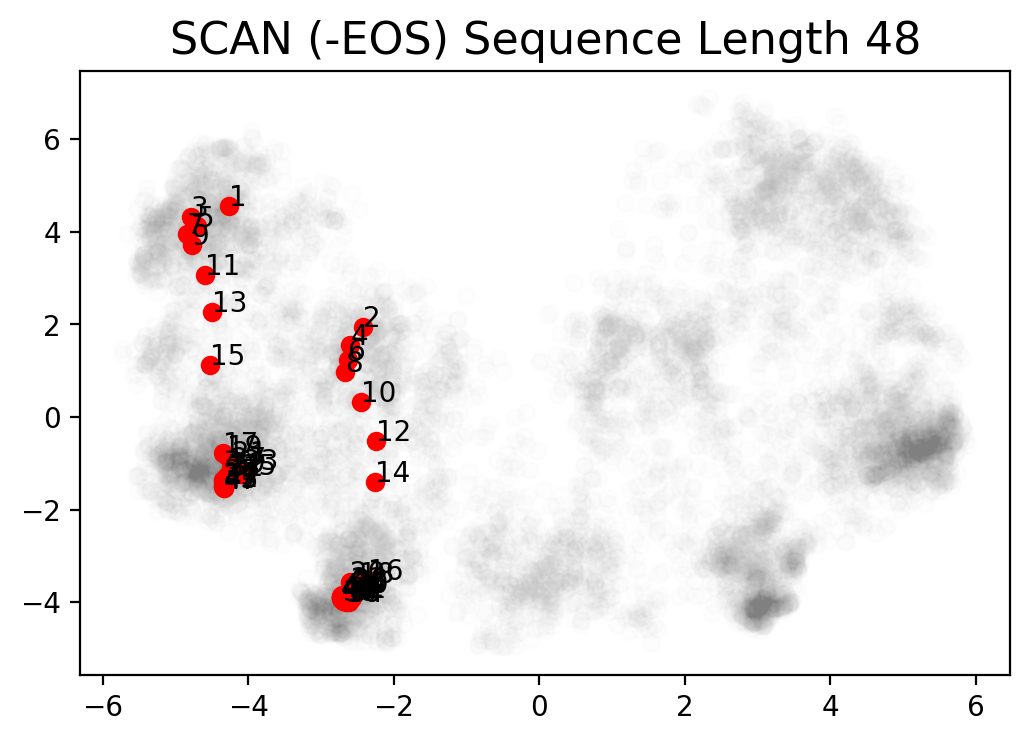}}
    \caption{Cherry-picked, but representative, examples of the paths of gold (i.e. not decoded) in-domain (length 4 and 16) and out-of-domain (length $24$, $33$, $48$) examples through the hidden state space of LSTMs trained on the SCAN $22$-token length split (top two principle components). Note that the \EOS token is included in the \EOSp plots, giving them one extra point. Interestingly, the hidden states from the OOD sequences arrive at the \EOS attractor well after the maximum training length (22) (for example, the sequence of length 48 arrives at the attractor at the hidden state at position 33), suggesting that the \EOSp does have some extrapolative abilities, but the \EOS attractor is suppressing them.}
    \label{fig:scan-ex}
\end{figure*}

\begin{table*}
    \centering
    \small
    \begin{tabular}{p{3mm} r l | r l | r l}
        \toprule
         & \multicolumn{2}{c}{\makecell{run around left after \\ jump around left twice}} & \multicolumn{2}{c}{look around left thrice} & \multicolumn{2}{c}{\makecell{run around left thrice after\\jump around right thrice}}\\
         \midrule
          & \EOSp & \EOSm & \EOSp & \EOSm & \EOSp & \EOSm  \\
1 & TURN\_LEFT & TURN\_LEFT & TURN\_LEFT & TURN\_LEFT & TURN\_RIGHT & TURN\_RIGHT \\
2 & JUMP & JUMP & LOOK & LOOK & JUMP & JUMP \\
3 & TURN\_LEFT & TURN\_LEFT & TURN\_LEFT & TURN\_LEFT & TURN\_RIGHT & TURN\_RIGHT \\
4 & JUMP & JUMP & LOOK & LOOK & JUMP & JUMP \\
5 & TURN\_LEFT & TURN\_LEFT & TURN\_LEFT & TURN\_LEFT & TURN\_RIGHT & TURN\_RIGHT \\
6 & JUMP & JUMP & LOOK & LOOK & JUMP & JUMP \\
7 & TURN\_LEFT & TURN\_LEFT & TURN\_LEFT & TURN\_LEFT & TURN\_RIGHT & TURN\_RIGHT \\
8 & JUMP & JUMP & LOOK & LOOK & JUMP & JUMP \\
9 & TURN\_LEFT & TURN\_LEFT & TURN\_LEFT & TURN\_LEFT & TURN\_RIGHT & TURN\_RIGHT \\
10 & JUMP & JUMP & LOOK & LOOK & JUMP & JUMP \\
11 & TURN\_LEFT & TURN\_LEFT & TURN\_LEFT & TURN\_LEFT & TURN\_RIGHT & TURN\_RIGHT \\
12 & JUMP & JUMP & LOOK & LOOK & JUMP & JUMP \\
13 & TURN\_LEFT & TURN\_LEFT & TURN\_LEFT & TURN\_LEFT & TURN\_RIGHT & TURN\_RIGHT \\
14 & JUMP & JUMP & LOOK & LOOK & JUMP & JUMP \\
15 & TURN\_LEFT & TURN\_LEFT & TURN\_LEFT & TURN\_LEFT & TURN\_RIGHT & TURN\_RIGHT \\
16 & JUMP & JUMP & LOOK & LOOK & JUMP & JUMP \\
17 & TURN\_LEFT & TURN\_LEFT & \color{red}LOOK & \color{blue}TURN\_LEFT & \color{blue} TURN\_RIGHT & \color{red} TURN\_LEFT \\
18 & \color{red}JUMP & \color{blue}RUN & \color{red}JUMP & \color{blue}LOOK & \color{blue} JUMP & \color{red}RUN \\
19 & TURN\_LEFT & TURN\_LEFT & \color{red}LOOK & \color{blue}TURN\_LEFT & \color{red}TURN\_LEFT & \color{red}TURN\_LEFT \\
20 & RUN & RUN & \color{red}JUMP & \color{blue}LOOK & \color{red}RUN & \color{red}RUN \\
21 & TURN\_LEFT & TURN\_LEFT & \color{red}LOOK & \color{blue}TURN\_LEFT & \color{red}TURN\_LEFT & \color{red}TURN\_LEFT \\
22 & RUN & RUN & \color{red}JUMP & \color{blue}LOOK & \color{red}RUN & \color{red}RUN \\
23 & TURN\_LEFT & TURN\_LEFT & \color{red}LOOK & \color{blue}TURN\_LEFT & \color{red}TURN\_LEFT & \color{red}TURN\_LEFT \\
24  & RUN & RUN & LOOK & LOOK &             \color{red}RUN & \color{red}RUN \\
25  &     &     &      &      &             TURN\_LEFT & TURN\_LEFT \\
26  &     &     &      &      &             RUN & RUN \\
27  &     &     &      &      &             TURN\_LEFT & TURN\_LEFT \\
28  &     &     &      &      &             RUN & RUN \\
29  &     &     &      &      &             TURN\_LEFT & TURN\_LEFT \\
30  &     &     &      &      &             RUN & RUN \\
31  &     &     &      &      &             TURN\_LEFT & TURN\_LEFT \\
32  &     &     &      &      &             RUN & RUN \\
33  &     &     &      &      &             \color{red}RUN & \color{blue}TURN\_LEFT \\
34  &     &     &      &      &             RUN & RUN \\  
35  &     &     &      &      &             \color{red}RUN & \color{blue}TURN\_LEFT \\  
36  &     &     &      &      &             RUN & RUN \\  
37  &     &     &      &      &             \color{red}RUN & \color{blue}TURN\_LEFT \\  
38  &     &     &      &      &             RUN & RUN \\  
39  &     &     &      &      &             \color{red}RUN & \color{blue}TURN\_LEFT \\  
40  &     &     &      &      &             RUN & RUN \\  
41  &     &     &      &      &             \color{red}RUN & \color{blue}TURN\_LEFT \\  
42  &     &     &      &      &             RUN & RUN \\
43  &     &     &      &      &             \color{red}RUN & \color{blue}TURN\_LEFT \\  
44  &     &     &      &      &             RUN & RUN \\  
45  &     &     &      &      &             \color{red}RUN & \color{blue}TURN\_LEFT \\
46  &     &     &      &      &             RUN & RUN \\  
47  &     &     &      &      &             \color{red}RUN & \color{blue}TURN\_LEFT \\  
48  &     &     &      &      &             RUN & RUN \\ 
         \bottomrule
    \end{tabular}
    \caption{More illustrative selections of errors the \EOSp and \EOSm models make when trained on the SCAN 22-token length split. In the left-most column we see that the \EOSp model makes a mistake on the conjunction that the \EOSm model does not make. In the middle column we see the \EOSp model fail on the template that is not seen at training time. In the right-most column we see an example where both models fail: the conjunction of two templates not seen at training time. The \EOSp model begins the third ``jump around'' but does not finish it, while the \EOSm model switches completely to the first ``turn left''.}
\end{table*}

\end{document}